\documentclass[letterpaper, 10 pt, conference]{ieeeconf}  

\IEEEoverridecommandlockouts                              

\usepackage{graphicx} 
\usepackage{amsmath} 
\usepackage{amssymb}  
\usepackage{cleveref} 
\usepackage{booktabs} 
\usepackage{multirow} 
\usepackage[table]{xcolor} 
\usepackage{censor}
\usepackage{pifont}
\usepackage{tabularx}    
\usepackage{colortbl}    

\definecolor{first}{RGB}{197,229,207}   
\definecolor{second}{RGB}{230,238,194}  
\definecolor{third}{RGB}{254,248,202}   

\title{\LARGE \bf
Memory Over Maps: \\ 3D Object Localization Without Reconstruction
}

\author{
Rui Zhou$^{1*}$, Xander Yap$^{2*}$, Jianwen Cao$^{1}$,  Allison Lau$^{2}$, Boyang Sun$^{2}$, Marc Pollefeys$^{2,3}$\\
{\small $^{*}$Equal contribution \hspace{1.5em} $^{1}$University of Zurich \hspace{1.5em} $^{2}$ETH Zurich \hspace{1.5em} $^{3}$Microsoft}
}

\begin{document}

\maketitle
\thispagestyle{empty}
\pagestyle{empty}

\begin{abstract}

Target localization is a prerequisite for embodied tasks such as navigation and manipulation. 
Conventional approaches rely on constructing explicit 3D scene representations to enable target localization, such as point clouds, voxel grids, or scene graphs. 
While effective, these pipelines incur substantial mapping time, storage overhead, and scalability limitations. 
Recent advances in vision-language models suggest that rich semantic reasoning can be performed directly on 2D observations, raising a fundamental question: is a complete 3D scene reconstruction necessary for object localization?
In this work, we revisit object localization and propose a map-free pipeline that stores only posed RGB-D keyframes as a lightweight visual memory—without constructing any global 3D representation of the scene. 
At query time, our method retrieves candidate views, re-ranks them with a vision-language model, and constructs a sparse, on-demand 3D estimate of the queried target through depth backprojection and multi-view fusion. 
Compared to reconstruction-based pipelines, this design drastically reduces preprocessing cost, enabling scene indexing that is over two orders of magnitude faster to build while using substantially less storage.
We further validate the localized targets on downstream object-goal navigation tasks. 
Despite requiring no task-specific training, our approach achieves strong performance across multiple benchmarks, demonstrating that direct reasoning over image-based scene memory can effectively replace dense 3D reconstruction for object-centric robot navigation. Project page: https://ruizhou-cn.github.io/memory-over-maps/
\end{abstract}

\section{INTRODUCTION} \label{sec:intro}

Finding the correct target for interaction is a fundamental prerequisite for many autonomous robotic applications, such as navigation \cite{savva2019habitat}, object search \cite{gadre2023cows}, and manipulation \cite{ahn2022saycan}.
The success of robot autonomy relies on precise localization of the target in the environment \cite{takmaz2023openmask3d, arnaudmcvay2025locate3d}.
When the environment is completely unknown, the robot must jointly explore the space while reasoning about potential target locations \cite{chaplot2020semexp, yokoyama2024vlfm}. 
However, in many practical scenarios, the robot has prior access to the environment, for example through a pre-scan using onboard sensors. 
This setting is common in service robotics, vacuum cleaning, inspection, and logistics applications. 
In such cases, efficiently leveraging prior observations to identify the goal in 3D space becomes crucial.

Existing approaches typically construct an explicit 3D reconstruction to enable target localization. 
The core idea is to fuse prior sensory inputs—such as images or LiDAR scans—into a consistent metric 3D representation of the environment. 
This representation provides geometric structure that supports navigation and planning \cite{reijgwart2020voxgraph, oleynikova2017voxblox, hornung2013octomap}. 
To localize semantic targets, subsequent works incorporate higher-level information such as semantic masks into the 3D map \cite{rosinol2020kimera, chang2021kimeramulti}. 
More recently, lightweight abstractions such as 3D scene graphs have been proposed to enable structured reasoning and incremental updates \cite{rosinol2020dynamicscenegraphs, hughes2022hydra, maggio2024clio}. 

With the rapid development of vision foundation models, large language models (LLMs), and vision-language models (VLMs), recent works have further injected semantic and language-aligned features into 3D maps to support open-vocabulary and concept-level queries \cite{huang2023vlmaps, Werby-RSS-24, gu2024conceptgraphs, arnaudmcvay2025locate3d}.
Despite these advances, such approaches fundamentally depend on accurate and often dense 3D reconstruction as an intermediate representation, which incurs substantial computational cost and storage overhead \cite{Werby-RSS-24}. 
Meanwhile, purely image-based methods have demonstrated strong performance in related tasks, including image retrieval \cite{arandjelovic2016netvlad}, feature matching \cite{sarlin2020superglue}, monocular depth estimation \cite{lin2025depth3}, and semantic reasoning \cite{liu2023llava}. 
This leads to a fundamental question: do we still need dense 3D maps as an intermediate representation for target localization?

\begin{figure}[t]
    \centering
    \includegraphics[width=\columnwidth]{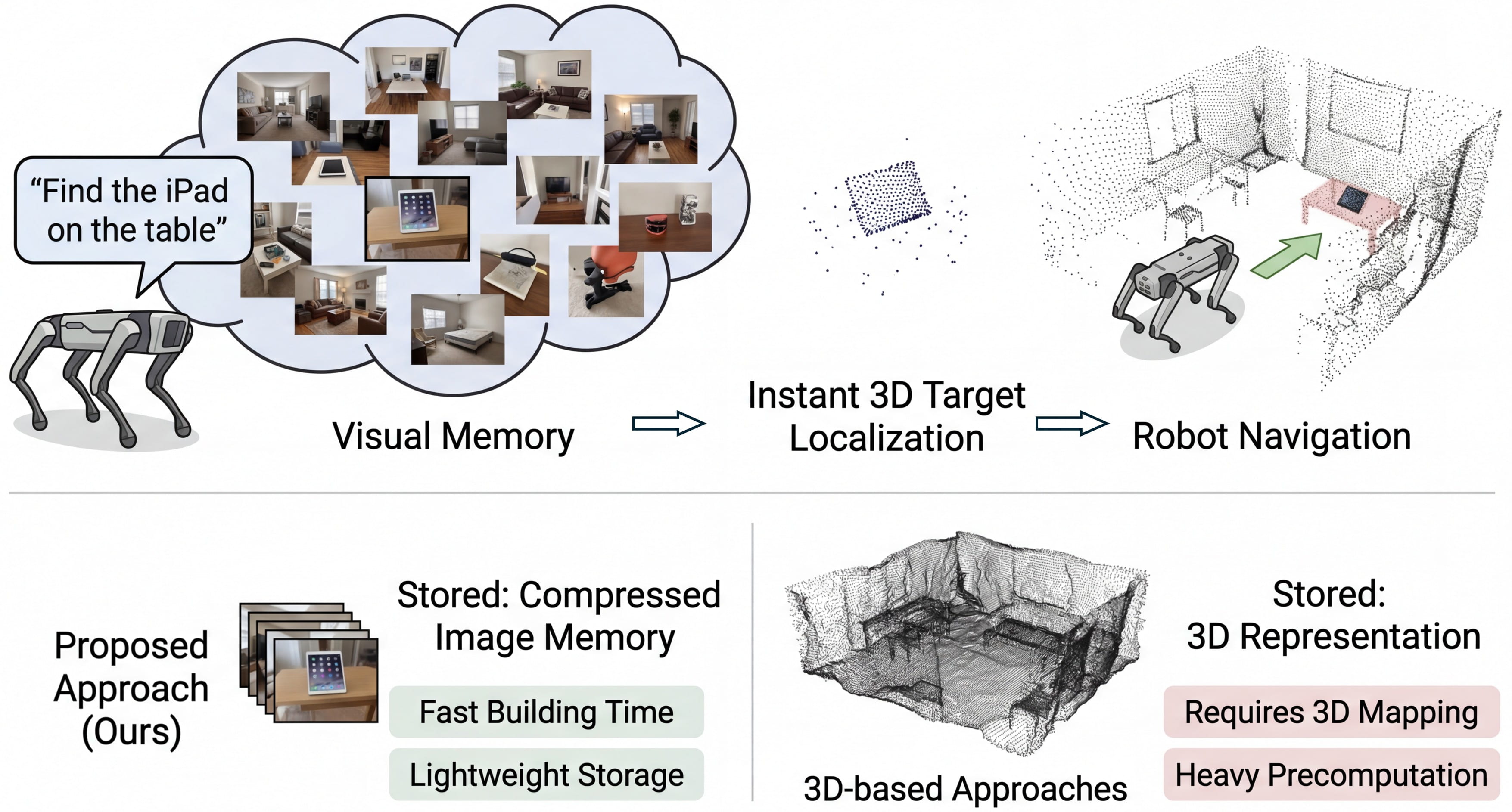}
    \vspace{-1.5em}
\caption{Image-based target localization without dense 3D reconstruction. Using only posed RGB-D keyframes as a lightweight visual memory, our method directly infers the 3D target location and enables a robot to navigate to the queried object without building a global scene map.}
    \label{fig:teaser}
    \vspace{-1.5em}
\end{figure}

\begin{figure*}[t]
    \centering
    \includegraphics[width=0.95\textwidth]{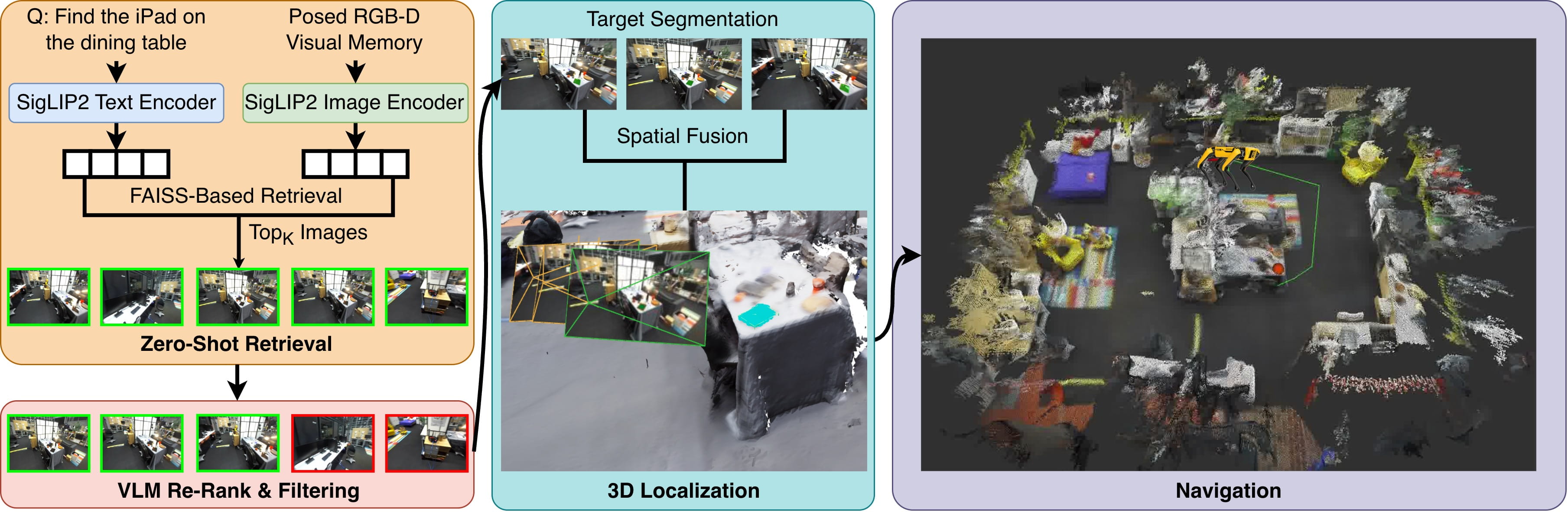}
    \caption{\textbf{Method overview.} Given a query and posed RGB-D keyframes: \textbf{(1)~Retrieval:} SigLIP2 embeddings indexed with FAISS retrieve top-$K$ candidates. \textbf{(2)~VLM Re-Rank:} a VLM filters false positives (red) and promotes true matches (green). \textbf{(3)~3D Localization:} SAM~3 segments the target; masked depth is backprojected, predictions are grouped into object instances, and per-instance multi-view fusion produces a 3D goal estimate. No global 3D map is built. \textbf{(4)~Navigation:} a PointNav policy navigates directly to the goal. (Scene point cloud shown for visualization only.)}
    \label{fig:method}
    \vspace{-1.5em}
\end{figure*}

Motivated by this question, we revisit the object localization problem. 
Given prior RGB-D scans of an environment, instead of constructing an explicit 3D reconstruction, we directly operate on the raw observational data, namely posed RGB-D images. Our pipeline treats posed RGB-D keyframes as a lightweight \emph{visual memory bank}. 
At query time, three stages convert a multimodal goal (language description, category label, or reference image) into a 3D target estimate: 
(1) SigLIP2~\cite{tschannen2025siglip2multilingualvisionlanguage} embeddings indexed with FAISS~\cite{johnson2019billion} retrieve candidate views in sub-second time; 
(2) a VLM re-ranks candidates by reasoning over fine-grained attributes; 
(3) text-prompted segmentation with depth backprojection and multi-view spatial fusion produces a robust 3D goal estimate. Our approach is training-free and avoids heavy preprocessing, making it both computationally and storage efficient. 
Despite its simplicity, it achieves strong performance in open-vocabulary object localization, particularly for small or fine-grained objects and context-dependent queries, where reconstruction-based approaches often struggle. 
These results suggest that eliminating explicit 3D intermediate representations is not only feasible but can improve localization performance. We further validate the localized goals in downstream object-goal navigation tasks and demonstrate strong, and in several settings state-of-the-art performance.
In addition, we deploy our pipeline in a real-world setting, where a ground robot searches for target objects using pre-scanned RGB-D sequences. As a summary, our contributions are:

\begin{itemize}
    \item A \textbf{map-free, open-vocabulary target localization framework} that operates directly on posed RGB-D keyframes through a two-stage hybrid retrieval pipeline, removing the need for explicit global 3D reconstruction and enabling scene indexing with over two orders of magnitude faster preprocessing.
    
    \item \textbf{Robust 3D localization and navigation}, combining text-prompted segmentation, depth backprojection, outlier removal, multi-view spatial fusion, and multi-goal fallback strategies for reliable target reaching.
    
    \item \textbf{Extensive evaluation} across multiple simulated benchmarks and real-world experiments, demonstrating strong localization performance and state-of-the-art downstream navigation performance without explicit 3D reconstruction.
\end{itemize}

\section{RELATED WORK}

\textbf{3D Reconstruction-Based Object Localization.}
The dominant paradigm builds an explicit 3D representation offline and queries it at inference time.
VLMaps~\cite{huang2023vlmaps} lifts LSeg~\cite{li2022language} features onto a bird's-eye-view grid; OpenMask3D~\cite{takmaz2023openmask3d} aggregates per-view CLIP features into reconstructed point clouds; Locate3D~\cite{arnaudmcvay2025locate3d} trains a 3D encoder to associate referring expressions with point-cloud regions.
Scene-graph methods~\cite{gu2024conceptgraphs,Werby-RSS-24,zhang2025openfungraph,werby2025keysg} introduce relational structure on top of reconstruction, while memory-based systems~\cite{liu2024dynamem,zhou2025lagmemo} maintain geometric point-cloud memories updated over time.
All these methods fundamentally depend on dense 3D reconstruction---incurring hours of preprocessing and gigabytes of per-scene storage~\cite{Werby-RSS-24}.

\textbf{2D Image-Based Retrieval and Grounding.}
Visual place recognition~\cite{arandjelovic2016netvlad,sarlin2020superglue} retrieves relevant images using learned descriptors but targets scene-level matching rather than object-level grounding.
Contrastive vision-language encoders---CLIP~\cite{radford2021clip}, ALIGN~\cite{jia2021align}, FLAVA~\cite{singh2022flava}, SigLIP2~\cite{tschannen2025siglip2multilingualvisionlanguage}. This enable zero-shot text-to-image retrieval but struggle with fine-grained attributes and spatial relationships.
VLMs~\cite{liu2023llava,Qwen2.5-VL} provide richer reasoning but are prohibitively expensive over large image collections.
Our method bridges these paradigms: fast embedding retrieval narrows thousands of keyframes to a handful of candidates, and a VLM re-ranks only those---combining scalability with reasoning depth, while avoiding any 3D reconstruction.

\textbf{Object Goal Navigation.}
Trained policies~\cite{wijmans2020ddppo,ramrakhya2022habitatweb,yadav2023ovrl,ramrakhya2023pirlnav} learn to navigate to object categories via reinforcement learning or imitation learning, with open-vocabulary extensions~\cite{majumdar2022zson,sun2024psl,chang2023goatthing,zhang2025uninavid} leveraging foundation models.
Zero-shot frontier methods~\cite{zhou2023esc,kuang2024openfmnav,yokoyama2024vlfm,yin2025unigoal,ziliotto2025tango,long2024instructnav,zhang2024trihelper,yin2024sgnav} bypass task-specific training but must search online, producing trajectories far longer than necessary; our approach replaces exploration with direct navigation to pre-localized targets.

\section{METHOD}
\label{sec:method}


\subsection{Overview}

Given a pre-scanned environment represented as an RGB-D video with known camera poses, and a multimodal query $q$ (natural-language description, category label, or reference image), our goal is to estimate the 3D location $\hat{\mathbf{p}} \in \mathbb{R}^3$ of the target object---without constructing any explicit 3D map of the scene.

We first extract a compact set of $N$ keyframes $\mathcal{K} = \{(I_n, D_n, P_n)\}_{n=1}^{N}$ from the raw video via pose-based selection (\Cref{subsec:scene_rep}), where $I_n$, $D_n$, and $P_n$ denote the RGB image, depth map, and camera-to-world pose, respectively. At query time, our pipeline proceeds in three stages:
\begin{enumerate}
    \item \textbf{Two-Stage Hybrid Retrieval} (\Cref{subsec:hybrid_retrieval}): a fast embedding search retrieves top-$K$ candidate views, followed by selective VLM re-ranking that filters false positives and scores fine-grained attributes.
    \item \textbf{3D Object Localization} (\Cref{subsec:3d_localization}): text-prompted segmentation with depth backprojection and multi-view spatial fusion produces a robust 3D goal estimate.
    \item \textbf{Goal-Directed Navigation} (\Cref{subsec:navigation}): a pre-trained PointNav policy navigates the agent directly to the localized goal, bypassing online exploration.
\end{enumerate}
The first two stages constitute a standalone localization system; the third extends it to embodied navigation.

\subsection{Scene Representation}
\label{subsec:scene_rep}

Given an RGB-D video sequence with known camera poses, we extract keyframes via pose-based selection: a new keyframe is accepted when rotation or translation relative to the last accepted frame exceeds a threshold ($\theta_{\text{rot}} = 15^\circ$, $\theta_{\text{trans}} = 0.25\text{m}$ for indoor scenes). This yields a compact database of $N$ keyframes $\mathcal{K} = \{(I_n, D_n, P_n)\}_{n=1}^{N}$. The representation can be further compressed by downsampling image resolution with minimal impact on retrieval accuracy (\Cref{subsec:compact_memory}).

\subsection{Two-Stage Hybrid Retrieval}
\label{subsec:hybrid_retrieval}


\textbf{Stage 1: Fast Candidate Retrieval.}
We encode each keyframe image $I_n$ into a visual embedding $\mathbf{v}_n = f(I_n) \in \mathbb{R}^d$ and each query into a query embedding $\mathbf{e}_q = g(q) \in \mathbb{R}^d$ using SigLIP2~\cite{tschannen2025siglip2multilingualvisionlanguage}, where $f$ and $g$ denote the image and text encoders, respectively; for image queries, $g$ is replaced by $f$. All embeddings are L2-normalized so that the inner product equals cosine similarity. We index $\{\mathbf{v}_n\}_{n=1}^{N}$ with FAISS~\cite{johnson2019billion} and retrieve the top-$K$ candidate set:
\begin{equation}
    \mathcal{C} = \operatorname*{arg\,top\text{-}K}_{n \in \{1, \ldots, N\}} \; \mathbf{e}_q^\top \mathbf{v}_n
\end{equation}
where $\mathcal{C} \subset \{1, \ldots, N\}$, $|\mathcal{C}| = K$. In the feature-only path, a deduplication step skips candidates whose cosine similarity to any already-selected result exceeds 0.9, avoiding redundant views from adjacent frames.

\textbf{Stage 2: VLM Re-ranking.}
Each of the $K$ candidates from Stage~1 is sent to a VLM with a structured prompt: \textit{``Is a \{query\} visible in this image? Reply ONLY: yes/no $\langle$visibility 0--10$\rangle$''}. This constrains output to ${\sim}$3--5 tokens (e.g., ``yes 8''). We normalize the returned integer score $s \in \{0, \ldots, 10\}$ to a confidence $c = s / 10$; candidates with negative detection are discarded and the rest re-ranked by $c$. If no candidate is confirmed, we fall back to the highest-confidence result from the VLM pass. Stage~1 reduces VLM queries from $N$ to $K$ via sub-millisecond vector search.

\subsection{3D Object Localization}
\label{subsec:3d_localization}

Given the top-$K$ retrieved views, we localize the target object in 3D through text-prompted segmentation, depth backprojection, multi-instance grouping, and per-instance multi-view spatial fusion.

\textbf{Text-Prompted Segmentation.}
For each retrieved view, we segment the target using SAM~3~\cite{carion2025sam3} with the query or category label for reference images as text prompt, selecting the highest-confidence mask. Views with median masked depth $>$4.0\,m are discarded, as distant views produce coarse masks that leak onto nearby surfaces, introducing large errors in depth backprojection.

\textbf{Depth Backprojection.}
Masked pixels within valid depth range (0.1--20.0\,m) are backprojected to 3D world coordinates via the pinhole camera model using known intrinsics and camera-to-world pose. The result is a sparse object point cloud $\mathcal{P}_n$ for each view.

\textbf{Multi-Instance Grouping.}
The top-$K$ predictions may correspond to multiple object instances. We group them by cloud proximity: each prediction is merged into an existing cluster if overlap ${\geq}$15\% or minimum point distance ${<}$0.5\,m; otherwise it seeds a new cluster. Each cluster yields a \emph{goal candidate}, sorted by L2 distance to the agent.

\textbf{Multi-View Spatial Fusion.}
A single view may only capture part of the target, producing an incomplete object cloud that biases the centroid and lacks surface points needed for close-range navigation. For each goal candidate, we recover a more complete representation by fusing multiple views. Starting from the seed cloud of the top-ranked view in the cluster, we verify whether other views in the same cluster observe the same instance by measuring 3D cloud overlap. If too few views confirm, we expand the search to nearby scene cameras whose frustum contains the target and repeat the segmentation--backprojection--overlap check. Confirmed views are fused by concatenating clouds. To remove ``flying pixels'' from depth discontinuities, background leakage from imprecise masks, and artifacts from imperfect mesh geometry, we cluster the merged cloud with HDBSCAN and retain only the largest cluster, preserving the original if it retains $<$5\% of points. The bounding box center of the fused cloud serves as the localization prediction, while the full cloud provides surface points for downstream navigation.

\subsection{Goal-Directed Navigation}
\label{subsec:navigation}

Given the 3D goal from \Cref{subsec:3d_localization}, we navigate directly using a learned PointNav policy, bypassing the online search that makes frontier exploration methods~\cite{yokoyama2024vlfm,yin2025unigoal} path-inefficient.

\textbf{PointNav Policy.}
We use a pre-trained DD-PPO~\cite{wijmans2020ddppo} PointNav checkpoint (VLFM~\cite{yokoyama2024vlfm} configuration): a ResNet-18 visual encoder processes $224{\times}224$ depth images, a goal embedding encodes relative polar coordinates $(\rho, \theta)$, and a 2-layer LSTM maintains state across steps, outputting discrete actions $\{\text{STOP}, \text{FORWARD}, \text{TURN\_LEFT}, \text{TURN\_RIGHT}\}$.

\textbf{Dynamic Goal Tracking.}
At each step, we project the object cloud into the current depth view and set the goal to the closest visible point, resetting the LSTM on goal updates.

\textbf{Multi-Goal Fallback.}
When multiple goal candidates exist, the agent navigates to the nearest instance first and switches to the next candidate if stuck.

\textbf{Visibility-Based Stopping.}
Before executing STOP, we project the object cloud into the current frame and verify against the depth buffer, overriding STOP with FORWARD if $<$5\% of in-view points are confirmed visible.

\section{EXPERIMENTS}

We compare against \emph{map-based methods} that invest in offline 3D reconstruction and \emph{zero-shot frontier exploration} methods that share our training-free property but rely on exhaustive online search.

\subsection{Experimental Setup}
\label{subsec:setup}

All experiments use SigLIP2~\cite{tschannen2025siglip2multilingualvisionlanguage} (SO400M) for feature extraction, FAISS~\cite{johnson2019billion} for indexing, SAM~3~\cite{carion2025sam3} for segmentation, and the DD-PPO PointNav policy from \Cref{subsec:navigation} (30$^\circ$ turn, 0.25\,m step, 79$^\circ$ HFOV, max 500 steps). VLM re-ranking uses Qwen2.5-VL-7B-Instruct~\cite{Qwen2.5-VL}. Defaults: $K{=}10$, $\theta_{\text{rot}}{=}15^\circ$, $\theta_{\text{trans}}{=}0.25$\,m, max 5 nearby views within 3\,m. All methods are provided with the same underlying pose information available in each benchmark; our method uses these poses only to store posed RGB-D keyframes and backproject localized targets, without constructing a global 3D map. All experiments were run on a single NVIDIA RTX 5090 GPU with 32 GB VRAM. For the localization benchmark (GOAT-Core), multi-view spatial fusion is applied independently to each of the top-$K$ predictions, which are then evaluated individually under the SR@K metric without instance grouping. For the navigation benchmarks, predictions are first grouped into distinct object instances via multi-instance grouping (\Cref{subsec:3d_localization}), then spatial fusion is applied per instance; the agent navigates to the nearest instance with multi-goal fallback. \colorbox{first}{Best}, \colorbox{second}{second-best}, and \colorbox{third}{third-best} results are highlighted accordingly.

\begin{table}[htbp]
\caption{\textbf{GOAT-Core localization.} SR@5 (\%) on GOAT-Core~\cite{zhou2025lagmemo}. ``--'' indicates the method does not support that query type.}
\label{tab:goatcore_loc}
\centering
\scriptsize\setlength{\tabcolsep}{3pt}
\begin{tabular*}{\columnwidth}{@{\extracolsep{\fill}}ll cccc}
\toprule
Scene & Method & Average & Object & Image & Language \\
\midrule

\multirow{6}{*}{\texttt{4ok}}
 & DynaMem~\cite{liu2024dynamem} & 10.1 & 13.2 & 17.1 & 0.0 \\
 & VLMaps~\cite{huang2023vlmaps}  & 63.3 & \cellcolor{third}75.5 & 41.8 & 70.8 \\
 & LagMemo~\cite{zhou2025lagmemo} & \cellcolor{second}86.7 & \cellcolor{first}\textbf{95.2} & \cellcolor{second}73.0 & \cellcolor{second}88.6 \\
 & OpenFunGraph~\cite{zhang2025openfungraph} & 77.5 & 73.7 & \cellcolor{third}70.7 & \cellcolor{third}87.8 \\
 & HOV-SG~\cite{Werby-RSS-24} & \cellcolor{third}78.5 & 73.7 & -- & 82.9 \\
 & Ours & \cellcolor{first}\textbf{90.1} & \cellcolor{second}92.1 & \cellcolor{first}\textbf{85.4} & \cellcolor{first}\textbf{92.7} \\

\midrule
\multirow{6}{*}{\texttt{5cd}}
 & DynaMem~\cite{liu2024dynamem} & 12.2 & 7.0 & 22.2 & 7.3 \\
 & VLMaps~\cite{huang2023vlmaps}  & 48.3 & 62.1 & \cellcolor{third}39.4 & 42.7 \\
 & LagMemo~\cite{zhou2025lagmemo} & \cellcolor{third}65.0 & \cellcolor{first}\textbf{80.2} & \cellcolor{second}63.1 & 49.6 \\
 & OpenFunGraph~\cite{zhang2025openfungraph} & 57.5 & \cellcolor{second}72.1 & 38.9 & \cellcolor{third}58.5 \\
 & HOV-SG~\cite{Werby-RSS-24} & \cellcolor{second}67.9 & 65.1 & -- & \cellcolor{second}70.7 \\
 & Ours & \cellcolor{first}\textbf{77.2} & \cellcolor{third}69.8 & \cellcolor{first}\textbf{86.1} & \cellcolor{first}\textbf{75.6} \\

\midrule
\multirow{6}{*}{\texttt{Nfv}}
 & DynaMem~\cite{liu2024dynamem} & 6.3 & 5.1 & 10.9 & 2.9 \\
 & VLMaps~\cite{huang2023vlmaps}  & \cellcolor{third}68.3 & \cellcolor{second}80.0 & \cellcolor{third}46.8 & \cellcolor{third}73.6 \\
 & LagMemo~\cite{zhou2025lagmemo} & 66.7 & \cellcolor{first}\textbf{85.5} & \cellcolor{second}49.6 & 61.1 \\
 & OpenFunGraph~\cite{zhang2025openfungraph} & 44.2 & 43.6 & 41.3 & 48.6 \\
 & HOV-SG~\cite{Werby-RSS-24} & \cellcolor{first}\textbf{82.4} & \cellcolor{third}76.9 & -- & \cellcolor{first}\textbf{88.6} \\
 & Ours & \cellcolor{second}71.6 & \cellcolor{third}76.9 & \cellcolor{first}\textbf{60.9} & \cellcolor{second}77.1 \\

\midrule
\multirow{6}{*}{\texttt{Tee}}
 & DynaMem~\cite{liu2024dynamem} & 10.2 & 4.9 & 15.8 & 9.8 \\
 & VLMaps~\cite{huang2023vlmaps}  & 55.0 & 61.3 & \cellcolor{third}45.1 & 56.9 \\
 & LagMemo~\cite{zhou2025lagmemo} & 65.0 & \cellcolor{first}\textbf{92.6} & 39.8 & 68.1 \\
 & OpenFunGraph~\cite{zhang2025openfungraph} & \cellcolor{third}73.3 & \cellcolor{third}78.1 & \cellcolor{second}63.2 & \cellcolor{third}78.1 \\
 & HOV-SG~\cite{Werby-RSS-24} & \cellcolor{second}78.1 & 68.3 & -- & \cellcolor{first}\textbf{87.8} \\
 & Ours & \cellcolor{first}\textbf{85.2} & \cellcolor{second}80.5 & \cellcolor{first}\textbf{92.1} & \cellcolor{second}82.9 \\

\midrule
\multirow{6}{*}{\textbf{Total}}
 & DynaMem~\cite{liu2024dynamem} & 9.6 & 7.5 & 16.1 & 5.1 \\
 & VLMaps~\cite{huang2023vlmaps}  & 58.8 & 69.7 & 43.3 & 61.0 \\
 & LagMemo~\cite{zhou2025lagmemo} & \cellcolor{third}70.8 & \cellcolor{first}\textbf{88.4} & \cellcolor{second}56.4 & 66.8 \\
 & OpenFunGraph~\cite{zhang2025openfungraph} & 63.1 & 67.1 & \cellcolor{third}53.4 & \cellcolor{third}69.0 \\
 & HOV-SG~\cite{Werby-RSS-24} & \cellcolor{second}76.5 & \cellcolor{third}70.8 & -- & \cellcolor{first}\textbf{82.3} \\
 & Ours & \cellcolor{first}\textbf{80.6} & \cellcolor{second}79.5 & \cellcolor{first}\textbf{80.1} & \cellcolor{first}\textbf{82.3} \\

\bottomrule
\end{tabular*}
\end{table}

\subsection{GOAT-Core 3D Localization}
\label{subsec:goatcore_loc}

We report \textbf{Success Rate at K (SR@K)}: an episode succeeds if any of the top-$K$ predictions falls within threshold $\tau$ of any ground-truth goal on the navigation plane. Let $E$ denote the number of episodes, $\hat{\mathbf{p}}_k \in \mathbb{R}^3$ the $k$-th predicted location, $\mathcal{G}_i$ the set of ground-truth goal positions for episode $i$, and $d_{\text{xz}}(\cdot,\cdot)$ the Euclidean distance on the xz-plane (ignoring height):

\begin{equation}
    \text{SR@K} = \frac{1}{E} \sum_{i=1}^{E} \mathbf{1}\!\left[\min_{k \leq K,\, \mathbf{g} \in \mathcal{G}_i} d_{\text{xz}}(\hat{\mathbf{p}}_k, \mathbf{g}) < \tau \right]
\end{equation}
We use $\tau = 1.5$\,m~\cite{zhou2025lagmemo} for all localization experiments.

\textbf{Task \& Setting.} Following prior work~\cite{zhou2025lagmemo}, we evaluate on GOAT-Core, a quality-controlled subset of GOAT-Bench~\cite{khanna2024goatbench} that addresses annotation issues in the full validation split. GOAT-Core selects four multi-room scenes (\texttt{4ok}, \texttt{5cd}, \texttt{Nfv}, \texttt{Tee}), averaging 7.25 rooms and 218.63\,m$^2$ per scene, with single-floor subtasks and manually corrected text descriptions. In total, GOAT-Core contains 480 multi-modal subtasks covering three query types: 158 \textit{object} (category-level), 163 \textit{image} (reference image), and 159 \textit{language} (natural-language description) goals. Image goals best stress-test cross-modal retrieval, while language goals exercise the fine-grained reasoning of VLM re-ranking. We report SR@5 per query type and the overall average (\Cref{tab:goatcore_loc}). We compare against VLMaps~\cite{huang2023vlmaps}, HOV-SG~\cite{Werby-RSS-24}, OpenFunGraph~\cite{zhang2025openfungraph}, DynaMem~\cite{liu2024dynamem}, and LagMemo~\cite{zhou2025lagmemo}.

\textbf{Results.} Our method achieves 82.3\% overall SR on language goals, matching HOV-SG (82.3\%) while requiring no 3D reconstruction. Our pipeline leads on image queries by a wide margin (80.1\% vs.\ 56.4\%), demonstrating that two-stage retrieval generalizes across modalities. On language queries requiring fine-grained reasoning (e.g., ``the red chair near the window''), we match or surpass reconstruction-based baselines. On object (category-level) queries, however, LagMemo retains a lead (88.4\% vs.\ 79.5\%): broad labels like ``chair'' match many views weakly, yielding noisier masks and less precise backprojection, whereas reconstruction-based methods provide pre-built 3D coordinates for every instance. Despite this gap, our method compensates on the more challenging image queries, achieving the highest average SR.

\begin{table}[htbp]
\caption{\textbf{Object goal navigation.} SR and SPL (\%) on (a)~HM3D~\cite{ramakrishnan2021hm3d} \& MP3D~\cite{chang2017matterport3d}, and (b)~HM3D-OVON~\cite{yokoyama2024hm3dovon}.}
\label{tab:objectnav}
\label{tab:hm3d_ovon}
\centering
\scriptsize\setlength{\tabcolsep}{3pt}

\textit{(a) HM3D and MP3D}\\[2pt]
\begin{tabular*}{\columnwidth}{@{\extracolsep{\fill}}l cc cc cc}
\toprule
\multirow{2}{*}[-3pt]{Method}
& \multirow{2}{*}[-3pt]{\shortstack{Training\\Free}}
& \multirow{2}{*}[-3pt]{\shortstack{Open\\Vocab}}
& \multicolumn{2}{c}{HM3D}
& \multicolumn{2}{c}{MP3D} \\
\cmidrule(lr){4-5} \cmidrule(lr){6-7}
& & & SR$\uparrow$ & SPL$\uparrow$ & SR$\uparrow$ & SPL$\uparrow$ \\
\midrule
\multicolumn{7}{l}{\textit{Trained Closed-Vocabulary}} \\
L3MVN~\cite{yu2023l3mvn} & $\times$ & $\times$ & 54.2 & 25.5 & 34.9 & 14.5 \\
Habitat-Web~\cite{ramrakhya2022habitatweb} & $\times$ & $\times$ & 57.6 & 23.8 &  24.2 & 5.9\\
OVRL~\cite{yadav2023ovrl} & $\times$ & $\times$ & 62.0 & 26.8 & 28.6 & 7.4 \\
PIRLNav-IL~\cite{ramrakhya2023pirlnav} & $\times$ & $\times$ & 64.1 & 27.1 & -- & -- \\
OVRL-v2~\cite{yadav2023ovrlv2} & $\times$ & $\times$ & 64.7 & 28.1 & -- & -- \\
PIRLNav-IL-RL~\cite{ramrakhya2023pirlnav} & $\times$ & $\times$ & 70.4 & 34.1 & -- & -- \\
\midrule
\multicolumn{7}{l}{\textit{Trained Open-Vocabulary}} \\
ZSON~\cite{majumdar2022zson} & $\times$ & $\checkmark$ & 25.5 & 12.6 & 15.3 & 4.8 \\
PSL~\cite{sun2024psl} & $\times$ & $\checkmark$ & 42.4 & 19.2 & -- & -- \\
GOAT~\cite{chang2023goatthing} & $\times$ & $\checkmark$ & 50.6 & 24.1 & -- & -- \\
Uni-NaVid~\cite{zhang2025uninavid} & $\times$ & $\checkmark$ & \cellcolor{second}73.7 & \cellcolor{third}37.1 & -- & -- \\
\midrule
\multicolumn{7}{l}{\textit{Zero-Shot Frontier Exploration}} \\
VLFM~\cite{yokoyama2024vlfm} & $\checkmark$ & $\checkmark$ & 52.5 & 30.4 & 36.4 & \cellcolor{third}17.5 \\
SG-Nav~\cite{yin2024sgnav} & $\checkmark$ & $\checkmark$ & 54.0 & 24.9 & \cellcolor{third}40.2 & 16.0 \\
UniGoal~\cite{yin2025unigoal} & $\checkmark$ & $\checkmark$ & 54.5 & 25.1 & \cellcolor{second}41.0 & 16.4 \\
OpenFMNav~\cite{kuang2024openfmnav} & $\checkmark$ & $\checkmark$ & 54.9 & 24.4 & 37.2 & 15.7 \\
TriHelper~\cite{zhang2024trihelper} & $\checkmark$ & $\checkmark$ & 56.5 & 25.3 & -- & -- \\
InstructNav~\cite{long2024instructnav} & $\checkmark$ & $\checkmark$ & 58.0 & 20.9 & -- & -- \\
\midrule

Ours (w/o VLM) & $\checkmark$ & $\checkmark$ & \cellcolor{third}73.5 & \cellcolor{second}48.2 & \cellcolor{first}\textbf{54.5} & \cellcolor{second}30.5 \\
Ours & $\checkmark$ & $\checkmark$ & \cellcolor{first}\textbf{76.7} & \cellcolor{first}\textbf{50.7} & \cellcolor{first}\textbf{54.5} & \cellcolor{first}\textbf{31.8} \\
\bottomrule
\end{tabular*}

\vspace{4pt}
\textit{(b) HM3D-OVON}\\[2pt]
\begin{tabular*}{\columnwidth}{@{\extracolsep{\fill}}l cc cc cc}
\toprule
\multirow{2}{*}{Method} & \multicolumn{2}{c}{\shortstack{Val\\Seen}} & \multicolumn{2}{c}{\shortstack{Val Seen\\Synonyms}} & \multicolumn{2}{c}{\shortstack{Val\\Unseen}} \\
\cmidrule(lr){2-3} \cmidrule(lr){4-5} \cmidrule(lr){6-7}
 & SR$\uparrow$ & SPL$\uparrow$ & SR$\uparrow$ & SPL$\uparrow$ & SR$\uparrow$ & SPL$\uparrow$ \\
\midrule

\multicolumn{7}{l}{\textit{Trained Closed-Vocabulary}} \\
BC~\cite{yokoyama2024hm3dovon} & 11.1 & 4.5 & 9.9 & 3.8 & 5.4 & 1.9 \\
DAgger~\cite{yokoyama2024hm3dovon} & 18.1 & 9.4 & 15.0 & 7.4 & 10.2 & 4.7 \\
BCRL~\cite{yokoyama2024hm3dovon} & 20.2 & 8.2 & 15.2 & 5.3 & 8.0 & 2.8 \\
RL~\cite{yokoyama2024hm3dovon} & 39.2 & 18.7 & 27.8 & 11.7 & 18.6 & 7.5 \\
DAgRL~\cite{yokoyama2024hm3dovon} & 41.3 & 21.2 & 29.4 & 14.4 & 18.3 & 7.9 \\

\midrule
\multicolumn{7}{l}{\textit{Trained Open-Vocabulary}} \\
DAgRL+OD~\cite{yokoyama2024hm3dovon} & 38.5 & 21.1 & 39.0 & 21.4 & 37.1 & \cellcolor{third}19.9 \\
Uni-NaVid~\cite{zhang2025uninavid} & 41.3 & 21.1 & 43.9 & \cellcolor{third}21.8 & 39.5 & 19.8 \\
MTU3D~\cite{zhu2025mtu3d} & \cellcolor{second}55.0 & \cellcolor{third}23.6 & \cellcolor{third}45.0 & 14.7 & \cellcolor{third}40.8 & 12.1 \\

\midrule
\multicolumn{7}{l}{\textit{Zero-Shot Frontier Exploration}} \\
VLFM~\cite{yokoyama2024vlfm} & 35.2 & 18.6 & 32.4 & 17.3 & 35.2 & 19.6 \\
TANGO~\cite{ziliotto2025tango} & -- & -- & -- & -- & 35.5 & 19.5 \\

\midrule
Ours (w/o VLM) & \cellcolor{third}53.3 & \cellcolor{second}37.7 & \cellcolor{second}54.3 & \cellcolor{second}38.3 & \cellcolor{second}48.1 & \cellcolor{second}34.7 \\
Ours & \cellcolor{first}\textbf{56.2} & \cellcolor{first}\textbf{40.3} & \cellcolor{first}\textbf{59.3} & \cellcolor{first}\textbf{43.0} & \cellcolor{first}\textbf{52.5} & \cellcolor{first}\textbf{38.2} \\

\bottomrule
\end{tabular*}
\end{table}

\subsection{Object Goal Navigation Benchmarks}
\label{subsec:objectnav}

We evaluate on three object goal navigation benchmarks using \texttt{habitat.Env}. An episode succeeds if the agent calls STOP within geodesic distance $\delta$ of a goal viewpoint ($\delta{=}0.1$\,m for HM3D/MP3D, $0.25$\,m for OVON). Let $S_i \in \{0,1\}$ be the success indicator for episode $i$, $l_i$ the agent's path length, and $l_i^*$ the shortest-path geodesic distance. We report \textbf{SR} $= \frac{1}{E} \sum_{i} S_i$ and \textbf{SPL}~\cite{anderson2018spl} $= \frac{1}{E} \sum_{i} S_i \cdot l_i^* / \max(l_i, l_i^*)$, which penalizes detours relative to the optimal path.

\subsubsection{HM3D Object Goal Navigation}
\label{subsec:hm3d_objectnav}

\textbf{Task \& Setting.} HM3D ObjectNav v2~\cite{ramakrishnan2021hm3d}: 1{,}000 episodes, 36 scenes, 6 categories, $\delta = 0.1$\,m. Baselines span trained closed-vocabulary~\cite{wijmans2020ddppo,chaplot2020semexp,ramrakhya2022habitatweb,yadav2023ovrl,ramrakhya2023pirlnav,yadav2023ovrlv2,chang2023goatthing,yu2023l3mvn}, trained open-vocabulary~\cite{majumdar2022zson,sun2024psl,zhang2025uninavid}, and zero-shot methods~\cite{zhou2023esc,kuang2024openfmnav,yokoyama2024vlfm,yin2025unigoal,long2024instructnav,zhang2024trihelper,yin2024sgnav} (\Cref{tab:objectnav}~a).

\textbf{Results.} Our pipeline achieves 76.7\% SR and 50.7\% SPL, the highest among all baselines. We exceed the best frontier method (VLFM, 30.4\% SPL) by +20.3\%, confirming that pre-localized goals eliminate the path waste of online exploration. VLM re-ranking adds +3.2\% SR / +2.5\% SPL, with gains concentrated on visually ambiguous categories (e.g., plant, tv\_monitor). Even without the VLM, the embedding-only variant almost surpasses all prior methods in SPL.

\subsubsection{MP3D Object Goal Navigation}
\label{subsec:mp3d_objectnav}

\textbf{Task \& Setting.} We evaluate cross-dataset generalization on MP3D object goal navigation~\cite{chang2017matterport3d,savva2019habitat} (2{,}195 episodes, 11 scenes, 21 categories, $\delta = 0.1$\,m). Results are reported alongside HM3D in \Cref{tab:objectnav} (a).

\textbf{Results.} Our method achieves 54.5\% SR and 31.8\% SPL, outperforming UniGoal by +13.5\% SR and $1.9\times$ SPL without any MP3D-specific tuning. The embedding-only variant (54.5\% SR) is on par with the VLM version, consistent with the GOAT-Core finding that VLM re-ranking is less critical for broad category-level queries.

\subsubsection{HM3D-OVON Object Goal Navigation}
\label{subsec:hm3d_ovon}

\textbf{Task \& Setting.} HM3D-OVON~\cite{yokoyama2024hm3dovon} extends HM3D object goal navigation to 379 categories across three splits: \textit{val\_seen} (3{,}000 episodes), \textit{val\_seen\_synonyms} (3{,}000), and \textit{val\_unseen} (3{,}000), using the same 36 scenes with $\delta = 0.25$\,m. Baselines include trained closed/open-vocabulary~\cite{yokoyama2024hm3dovon,zhang2025uninavid,zhu2025mtu3d} and zero-shot methods~\cite{yokoyama2024vlfm,ziliotto2025tango} (\Cref{tab:objectnav} (b)).

\textbf{Results.} Our method achieves the best SR and SPL on all three splits. The advantage is most pronounced on SPL: on \textit{val\_unseen}, we reach 38.2\% vs.\ MTU3D's 12.1\% (${\sim}3\times$), indicating that our direct navigation generalizes to novel categories far better than trained policies. Performance is consistent across splits---unlike trained baselines that degrade sharply from \textit{val\_seen} to \textit{val\_unseen}---highlighting the open-vocabulary robustness of embedding-based retrieval.

\begin{table}[htbp]
\caption{\textbf{Text-to-image retrieval.} AR@1 and AR@5 on SUN RGB-D~\cite{song2015sunrgbd}.}
\label{tab:sunrgbd}
\centering
\scriptsize\setlength{\tabcolsep}{3pt}
\begin{tabular*}{\columnwidth}{@{\extracolsep{\fill}}l ccccc}
\toprule
Sensor & FLAVA~\cite{singh2022flava} & CLIP-B~\cite{radford2021clip} & ALIGN~\cite{jia2021align} & CLIP-L~\cite{radford2021clip} & Ours \\
\midrule
\multicolumn{6}{l}{\textit{AR@1$\uparrow$}} \\
KV1       & 5.3  & 60.0 & \cellcolor{second}67.5 & \cellcolor{third}62.0 & \cellcolor{first}\textbf{70.5} \\
KV2       & 26.2 & 48.4 & \cellcolor{third}64.5 & \cellcolor{second}69.8 & \cellcolor{first}\textbf{82.2} \\
Xtion     & 15.2 & 49.7 & \cellcolor{third}58.2 & \cellcolor{second}64.7 & \cellcolor{first}\textbf{73.1} \\
RealSense & 19.6 & \cellcolor{first}\textbf{66.7} & 46.6 & \cellcolor{third}60.2 & \cellcolor{second}62.0 \\
Avg.      & 16.6 & 56.2 & \cellcolor{third}59.2 & \cellcolor{second}64.2 & \cellcolor{first}\textbf{71.9} \\
\midrule
\multicolumn{6}{l}{\textit{AR@5$\uparrow$}} \\
KV1       & 38.4 & \cellcolor{third}83.9 & \cellcolor{second}86.0 & 83.1 & \cellcolor{first}\textbf{89.1} \\
KV2       & 30.7 & 78.4 & \cellcolor{third}80.0 & \cellcolor{second}84.1 & \cellcolor{first}\textbf{93.0} \\
Xtion     & 31.8 & 80.3 & \cellcolor{second}84.9 & \cellcolor{second}84.9 & \cellcolor{first}\textbf{89.9} \\
RealSense & 52.2 & \cellcolor{third}87.3 & 85.7 & \cellcolor{first}\textbf{90.0} & \cellcolor{second}88.5 \\
Avg.      & 38.3 & 82.5 & \cellcolor{third}84.2 & \cellcolor{second}85.5 & \cellcolor{first}\textbf{90.1} \\
\bottomrule
\end{tabular*}
\end{table}

\subsection{Image Retrieval Benchmarks}
\label{subsec:retrieval}

\textbf{Task \& Setting.} We evaluate text-to-image retrieval on SUN RGB-D~\cite{song2015sunrgbd} (${\sim}$10{,}335 indoor scenes, four sensor types). A separate FAISS index is built per sensor group, and we report per-instance AR@K---the fraction of queries where at least one relevant image appears in the top-$K$---against CLIP-base/large~\cite{radford2021clip}, ALIGN~\cite{jia2021align}, and FLAVA~\cite{singh2022flava} (\Cref{tab:sunrgbd}).

\textbf{Results.} Our method ranks first on three of four sensors (KV1, KV2, Xtion), with the largest gain on KV2 (+12.4\% AR@1 over CLIP-large). On RealSense, CLIP-base leads AR@1 (66.7\% vs.\ 62.0\%). Averaging across sensors, we achieve 71.9\% AR@1 and 90.1\% AR@5.

\subsection{Memory-Accuracy Trade-off and Efficiency}
\label{subsec:compact_memory}

\begin{table}[t]
\caption{\textbf{Memory-accuracy trade-off and pipeline efficiency.}}
\label{tab:memory_tradeoff}
\label{tab:efficiency}
\centering
\scriptsize\setlength{\tabcolsep}{3pt}

\textit{(a) HM3D ObjectNav SR and SPL (\%) under different resolutions and keyframe settings}\\[2pt]
\begin{tabular*}{\columnwidth}{@{\extracolsep{\fill}}lccccc}
\toprule
Setting & Resolution & \#Frames & Storage$\downarrow$ & SR$\uparrow$ & SPL$\uparrow$ \\
\midrule
Full frames & $1600{\times}1200$ & 127177   & 150.1\,GB & \cellcolor{third}74.5 & 47.1 \\
Keyframes   & $1600{\times}1200$ & 36888   & \cellcolor{third}43.8\,GB  & \cellcolor{first}\textbf{76.7} & \cellcolor{first}\textbf{50.7} \\
Keyframes   & $640{\times}480$   & 36888   & \cellcolor{second}9.5\,GB   & \cellcolor{second}75.1 & \cellcolor{second}48.8 \\
Keyframes   & $160{\times}120$   & 36888   & \cellcolor{first}\textbf{0.9\,GB}   & 72.8 & \cellcolor{third}47.5 \\
\bottomrule
\end{tabular*}

\vspace{4pt}
\textit{(b) Build time, storage (summed over 4 GOAT-Core scenes~\cite{zhou2025lagmemo}), and per-query latency}\\[2pt]
\begin{tabular*}{\columnwidth}{@{\extracolsep{\fill}}l ccc ccc c}
\toprule
Method & Build$\downarrow$ & Store$\downarrow$ & Obj. & Img. & Lang. & Avg.Q$\downarrow$ \\
\midrule
VLMaps~\cite{huang2023vlmaps}  & \cellcolor{third}25.9\,min & \cellcolor{third}4.1\,GB & \cellcolor{second}55ms & \cellcolor{second}60ms & \cellcolor{second}58ms & \cellcolor{second}58\,ms \\
DynaMem~\cite{liu2024dynamem} & \cellcolor{second}19.2\,min & 18.8\,GB & \cellcolor{third}603ms & \cellcolor{third}324ms & \cellcolor{third}592ms & \cellcolor{third}506\,ms \\
HOV-SG~\cite{Werby-RSS-24}  & 5.3\,h & 34.3\,GB & 2.38s & -- & 2.40s & 2.39\,s \\
OpenFunGraph~\cite{zhang2025openfungraph}     & 2.3\,h & \cellcolor{second}2.86\,GB & \cellcolor{first}\textbf{25ms} & \cellcolor{first}\textbf{39ms} & \cellcolor{first}\textbf{25ms} & \cellcolor{first}\textbf{30\,ms} \\
\textbf{Ours} & \cellcolor{first}\textbf{35.5\,s} & \cellcolor{first}\textbf{1.4\,GB} & 3.15s & 2.86s & 2.74s & 2.92\,s \\
\bottomrule
\end{tabular*}
\end{table}

\textbf{Results.} \Cref{tab:memory_tradeoff}~a and~b summarize the trade-offs. Keyframe selection reduces the frame count from 127177 to 36888 (${\sim}$71\% fewer frames) while improving both SR (74.5\%$\to$76.7\%) and SPL (47.1\%$\to$50.7\%) by sharpening the embedding index and removing redundant views. Downsampling to 640$\times$480 incurs only $-$1.8\% SR and $-$2.0\% SPL while reducing storage ${\sim}$5$\times$ (43.8\,GB$\to$9.5\,GB); even at 160$\times$120, SR remains 72.8\% and SPL 47.5\% with just 0.9\,GB (0.6\% of original storage).

All baselines require explicit 3D reconstruction: VLMaps~\cite{huang2023vlmaps} projects LSeg~\cite{li2022language} features onto a bird's-eye-view grid (4.1\,GB); DynaMem~\cite{liu2024dynamem} and HOV-SG~\cite{Werby-RSS-24} store dense per-point features (18.8\,GB and 34.3\,GB); OpenFunGraph~\cite{zhang2025openfungraph} compresses data into object-instance nodes (2.86\,GB). Our method stores only posed RGB-D keyframes and SigLIP2 embeddings indexed with FAISS, yielding the smallest storage (1.4\,GB) and fastest build time (35.5\,s vs.\ 19.2\,min for the next-fastest baseline, a $32\times$ speedup). Per-query latency (2.92\,s) is dominated by local VLM inference and segmentation.

\begin{figure}[htbp]
    \centering
    \setlength{\tabcolsep}{1.5pt}
    \scriptsize
    \newcommand{\qualimg}[1]{\raisebox{-0.5\height}{\includegraphics[width=0.22\columnwidth]{#1}}}
    \begin{tabular}{@{}c cccc@{}}
    \rotatebox[origin=c]{90}{\textsc{HM3D}} &
    \qualimg{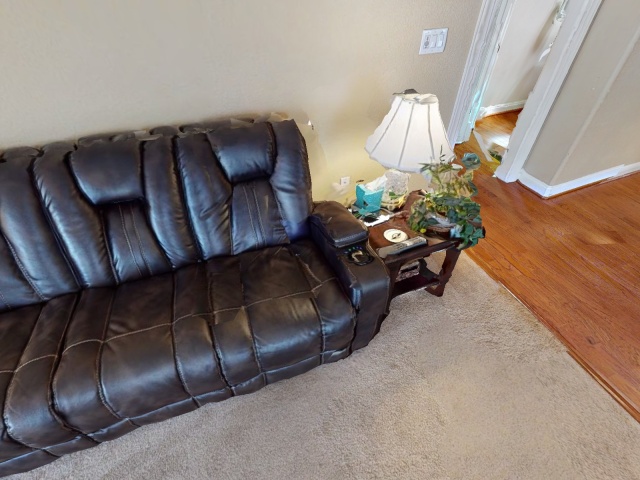} &
    \qualimg{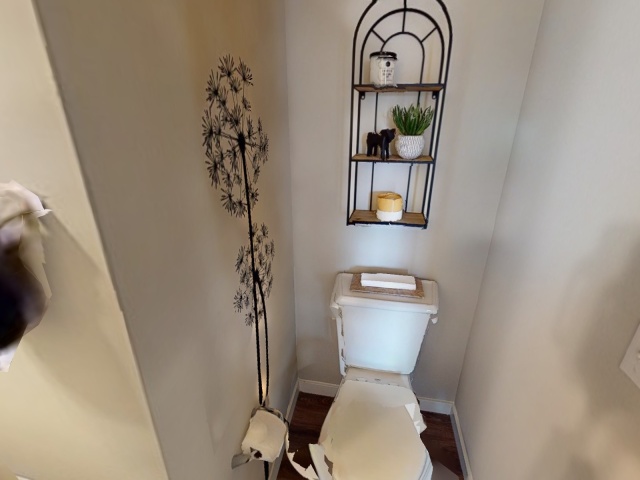} &
    \qualimg{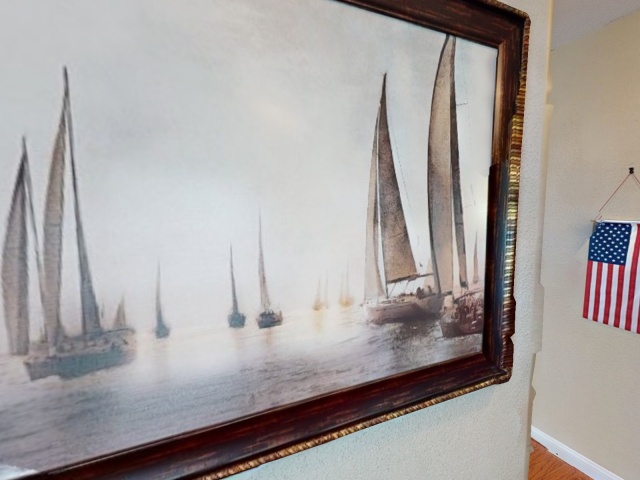} &
    \qualimg{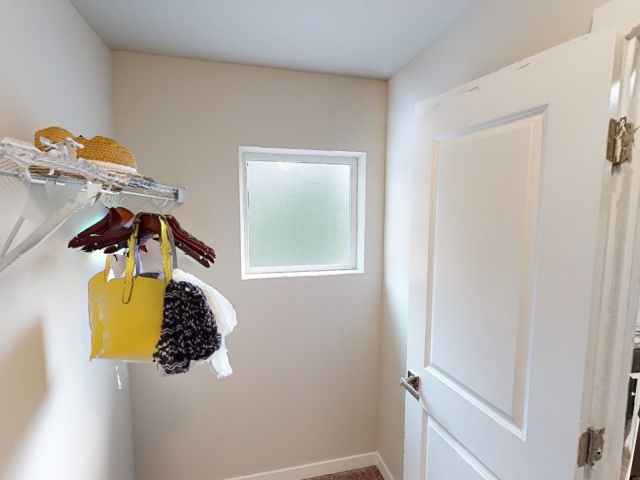} \\
    & \textit{Plant by sofa} & \textit{Plant above toilet} & \textit{Sailboat painting} & \textit{Bag in closet} \\[3pt]
    \rotatebox[origin=c]{90}{\textsc{MP3D}} &
    \qualimg{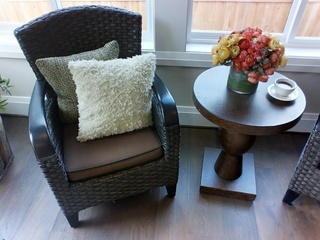} &
    \qualimg{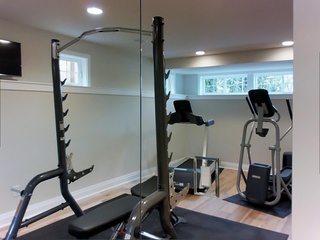} &
    \qualimg{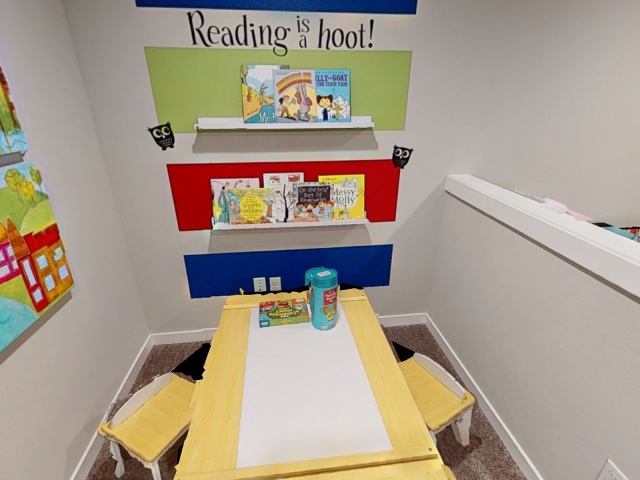} &
    \qualimg{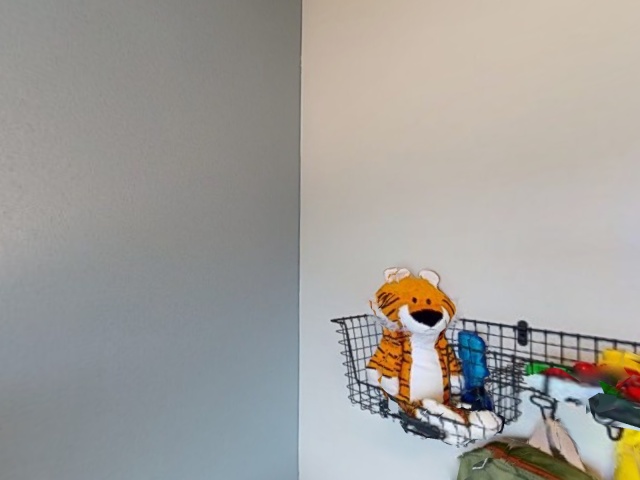} \\
    & \textit{Coffee by flower} & \textit{Place to workout} & \textit{Place to sit} & \textit{Tiger plushie} \\[3pt]
    \rotatebox[origin=c]{90}{\textsc{CAB}} &
    \qualimg{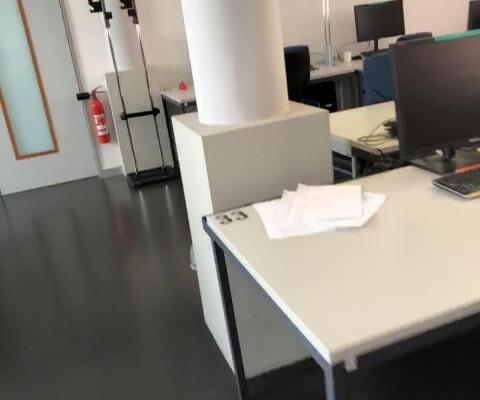} &
    \qualimg{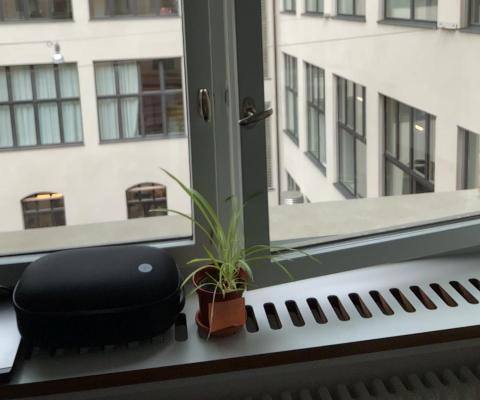} &
    \qualimg{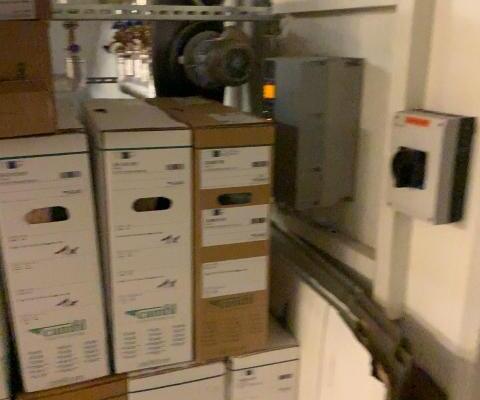} &
    \qualimg{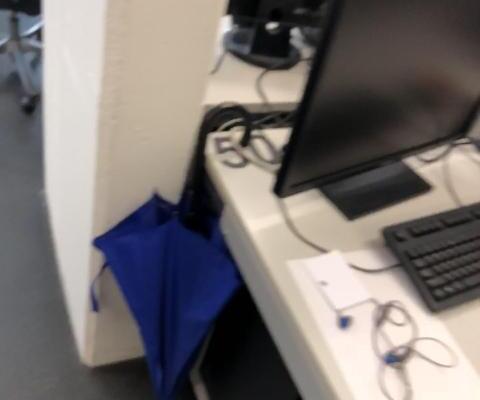} \\
    & \textit{Papers on desk} & \textit{Plant by window} & \textit{Light switch} & \textit{Earphone on desk} \\[3pt]
    \rotatebox[origin=c]{90}{\textsc{LIN}} &
    \qualimg{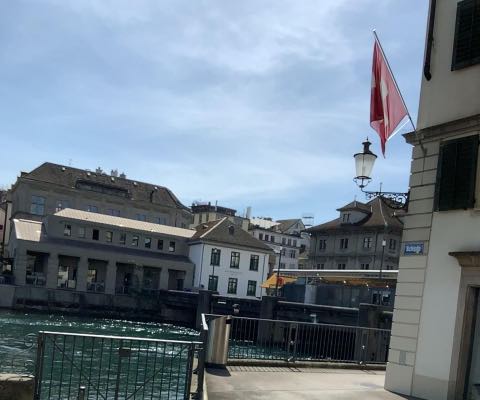} &
    \qualimg{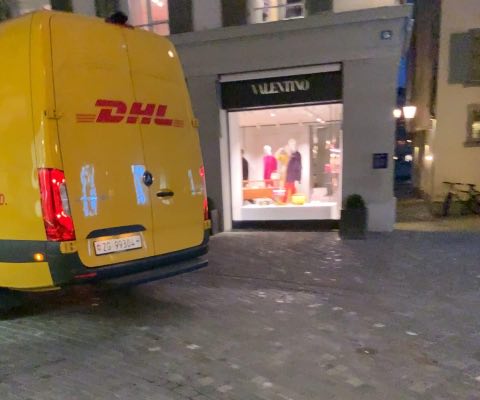} &
    \qualimg{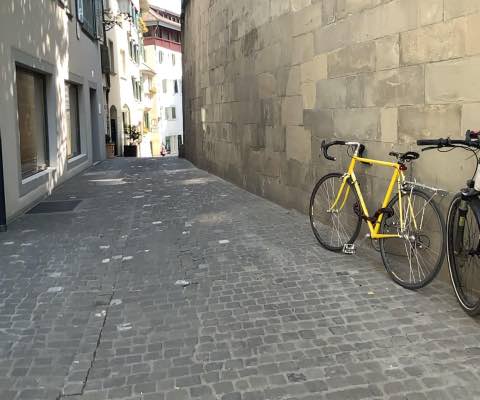} &
    \qualimg{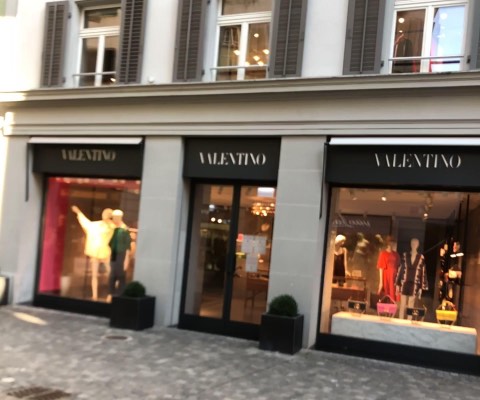} \\
    & \textit{Swiss flag by river} & \textit{DHL car} & \textit{Yellow bike} & \textit{Valentino store} \\
    \end{tabular}
    \caption{\textbf{Qualitative retrieval examples.} Top-1 retrieved images for natural-language queries across small indoor (HM3D~\cite{ramakrishnan2021hm3d}, MP3D~\cite{chang2017matterport3d}), large indoor (LaMAR-CAB~\cite{sarlin2022lamar}), and outdoor (LaMAR-LIN~\cite{sarlin2022lamar}) scenes.}
    \label{fig:simulationqualitative}
    \label{fig:qualitativeexamples}
\end{figure}

\begin{figure}[htbp]
    \centering
    \small
    \begin{minipage}{0.48\linewidth}
        \centering \small\textsc{HOV-SG}
    \end{minipage} \hfill
    \begin{minipage}{0.48\linewidth}
        \centering \small\textsc{Ours}
    \end{minipage}

    \vspace{0.3em}

    \begin{minipage}{0.48\linewidth}
        \begin{minipage}{0.48\linewidth}
            \includegraphics[width=\linewidth]{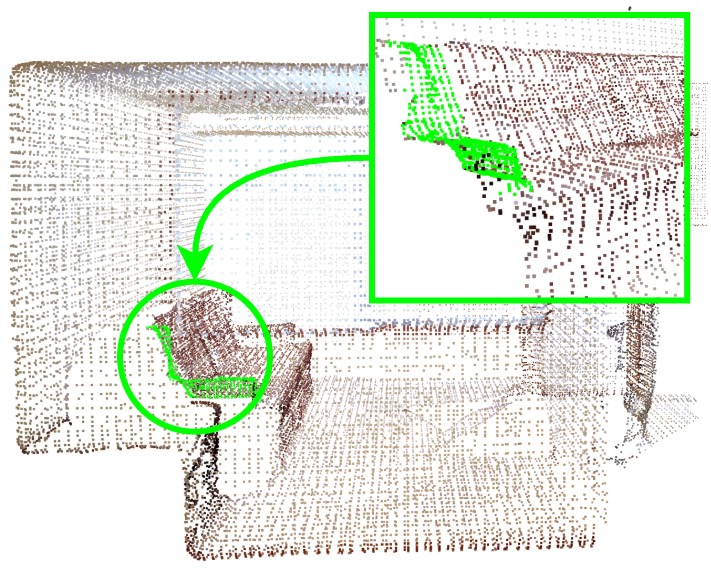}
        \end{minipage} \hfill
        \begin{minipage}{0.48\linewidth}
            \includegraphics[width=\linewidth]{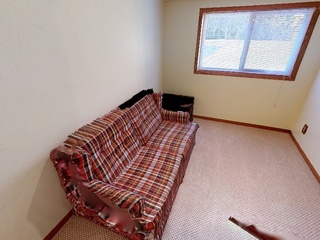}
        \end{minipage} \\[3pt]
        \centering \scriptsize\textit{Couch} ~{\color{green}\ding{51}}
    \end{minipage} \hfill
    \begin{minipage}{0.48\linewidth}
        \begin{minipage}{0.48\linewidth}
            \includegraphics[width=\linewidth]{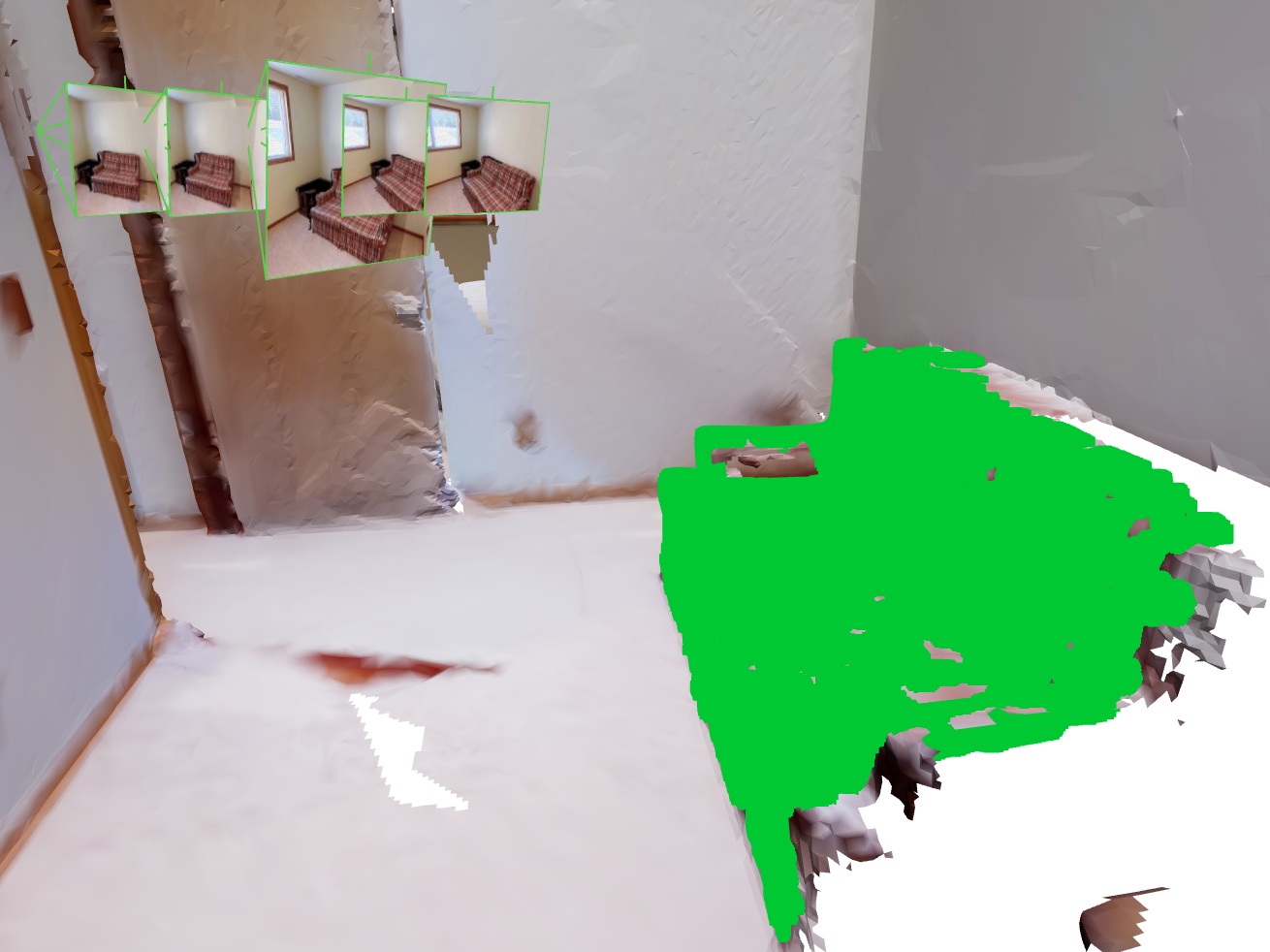}
        \end{minipage} \hfill
        \begin{minipage}{0.48\linewidth}
            \includegraphics[width=\linewidth]{figures/hovsg/couchhovsg.jpg}
        \end{minipage} \\[3pt]
        \centering \scriptsize\textit{Couch} ~{\color{green}\ding{51}}
    \end{minipage}

    \vspace{0.5em}

    \begin{minipage}{0.48\linewidth}
        \begin{minipage}{0.48\linewidth}
            \includegraphics[width=\linewidth]{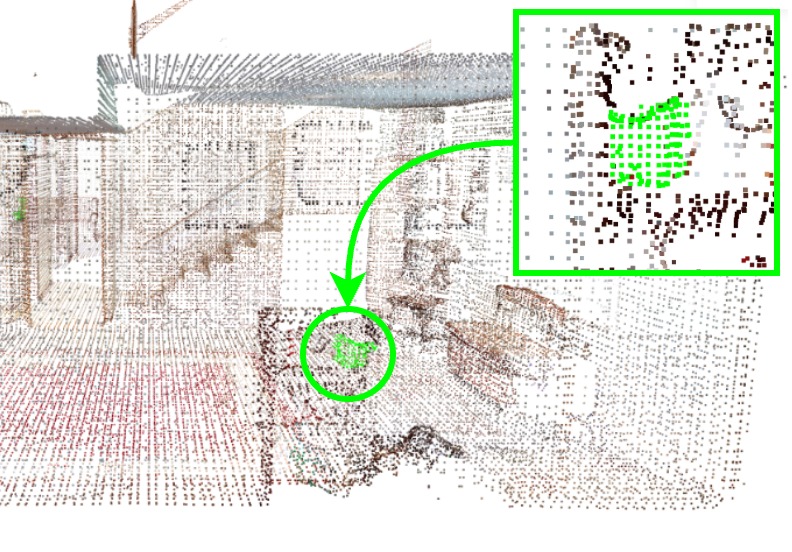}
        \end{minipage} \hfill
        \begin{minipage}{0.48\linewidth}
            \includegraphics[width=\linewidth]{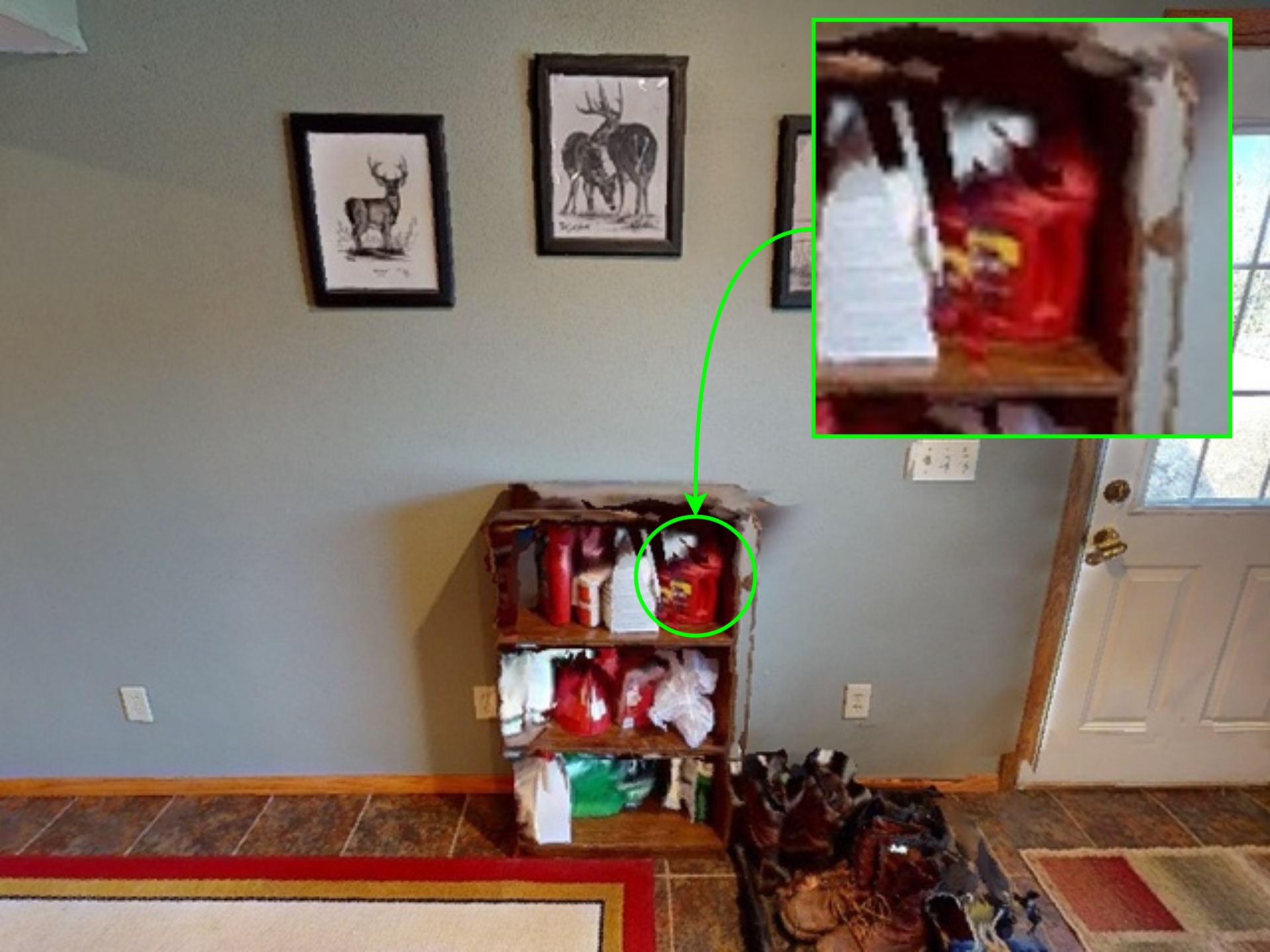}
        \end{minipage} \\[3pt]
        \centering \scriptsize\textit{Book on nightstand} ~{\color{red}\ding{55}}
    \end{minipage} \hfill
    \begin{minipage}{0.48\linewidth}
        \begin{minipage}{0.48\linewidth}
            \includegraphics[width=\linewidth]{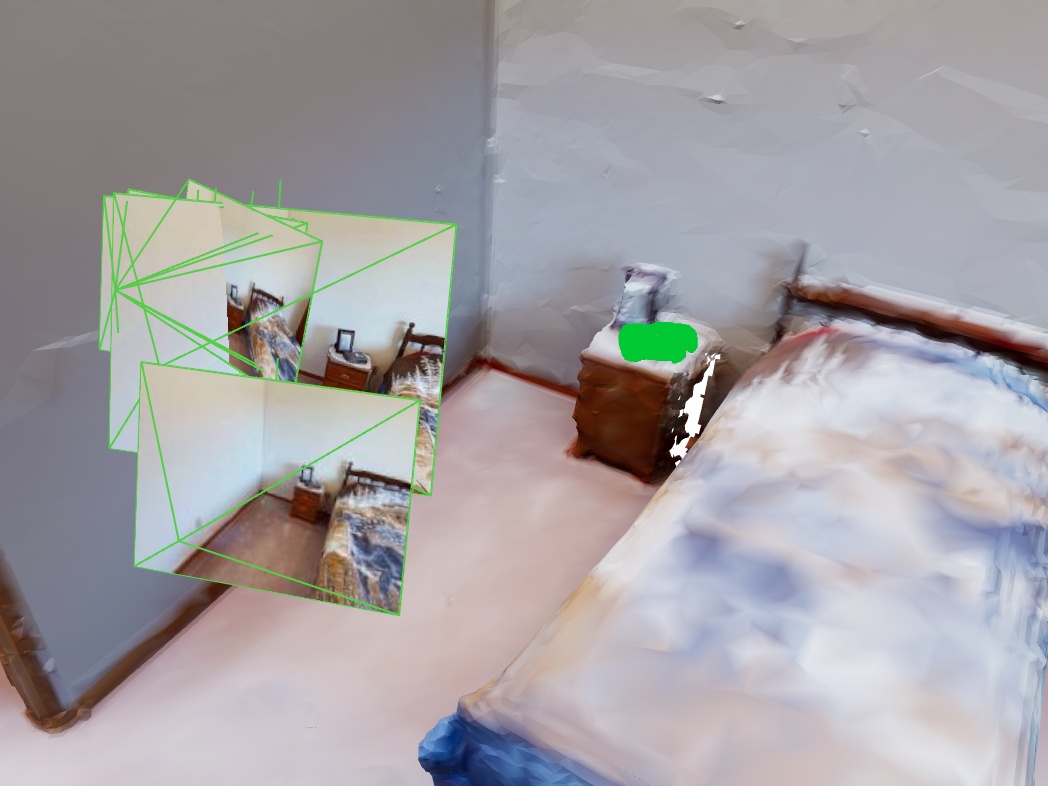}
        \end{minipage} \hfill
        \begin{minipage}{0.48\linewidth}
            \includegraphics[width=\linewidth]{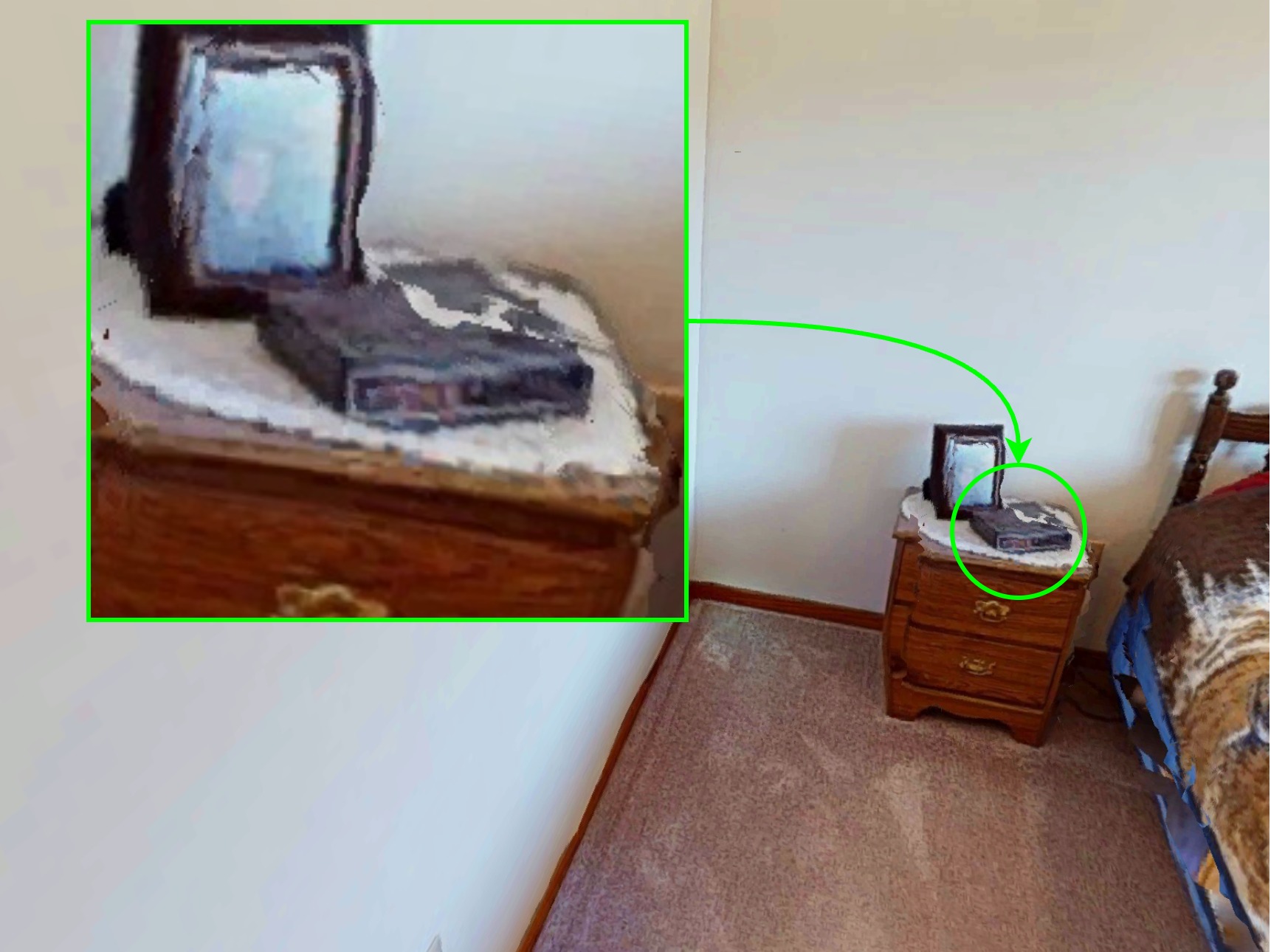}
        \end{minipage} \\[3pt]
        \centering \scriptsize\textit{Book on nightstand} ~{\color{green}\ding{51}}
    \end{minipage}

    \caption{\textbf{Qualitative comparison with HOV-SG~\cite{Werby-RSS-24}.} Our method retrieves the correct view with segmentation mask and produces a fused 3D point cloud, capturing fine-grained details that HOV-SG overlooks.}
    \label{fig:hovsgvsours_4col}
\end{figure}

\subsection{Qualitative Results}
\label{subsec:qualitative}

\Cref{fig:simulationqualitative} shows top-1 retrieved images for fine-grained and context-dependent queries across small indoor (HM3D, MP3D), large indoor (LaMAR-CAB~\cite{sarlin2022lamar}), and outdoor (LaMAR-LIN~\cite{sarlin2022lamar}) scenes, demonstrating retrieval across indoor and outdoor scenes of varying scale. \Cref{fig:hovsgvsours_4col} compares against HOV-SG~\cite{Werby-RSS-24}: scene-graph queries sometimes return incorrect regions (e.g., localizing detergent instead of a book), whereas our two-stage pipeline retrieves the correct view and produces a fused object point cloud for accurate 3D localization.

\begin{figure}[htbp]
    \centering
    \setlength{\tabcolsep}{1.5pt}
    \scriptsize
    \renewcommand{\arraystretch}{0.1}
    \begin{tabular}{@{}ccc@{}}
    \small\textsc{(a) Third-person} & \small\textsc{(b) Retrieved + mask} & \small\textsc{(c) Spot camera} \\[4pt]
    \includegraphics[width=0.32\columnwidth]{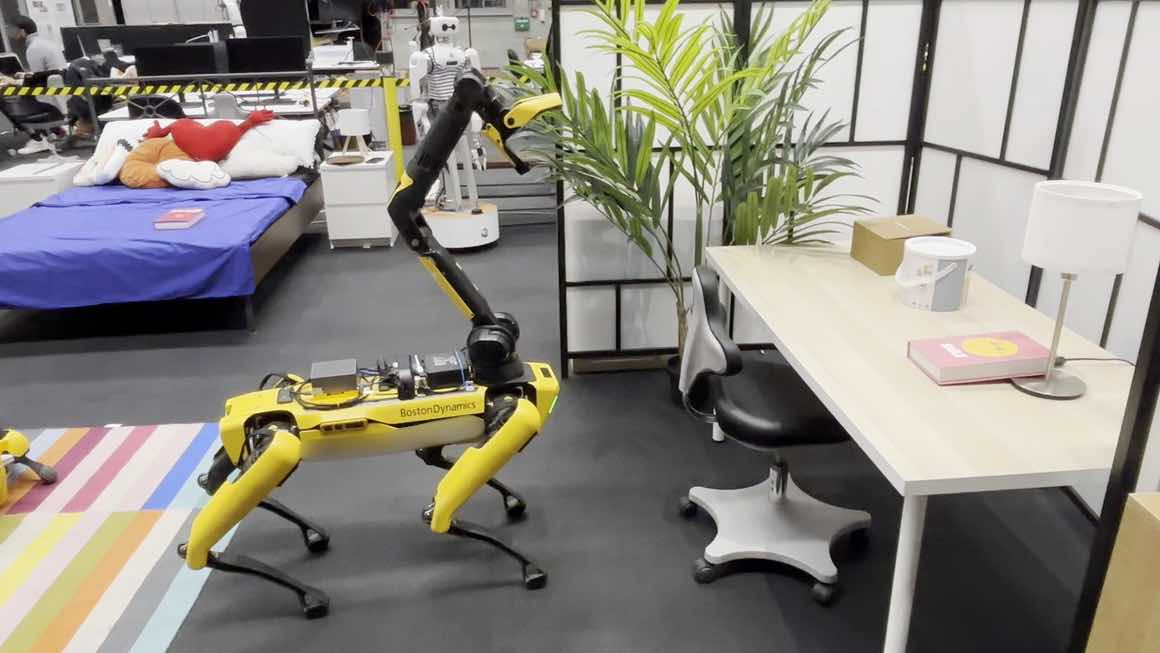} &
    \includegraphics[width=0.32\columnwidth]{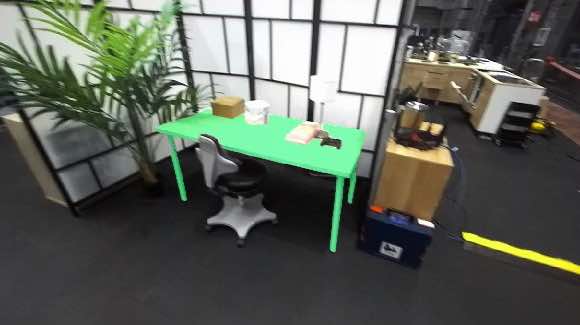} &
    \includegraphics[width=0.32\columnwidth]{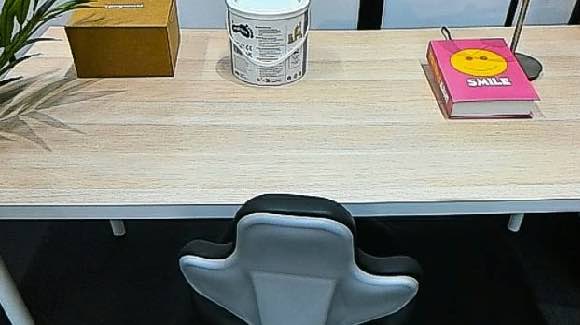} \\[1pt]
    \multicolumn{3}{@{}c@{}}{\scriptsize\textit{Table for one}} \\[3pt]
    \includegraphics[width=0.32\columnwidth]{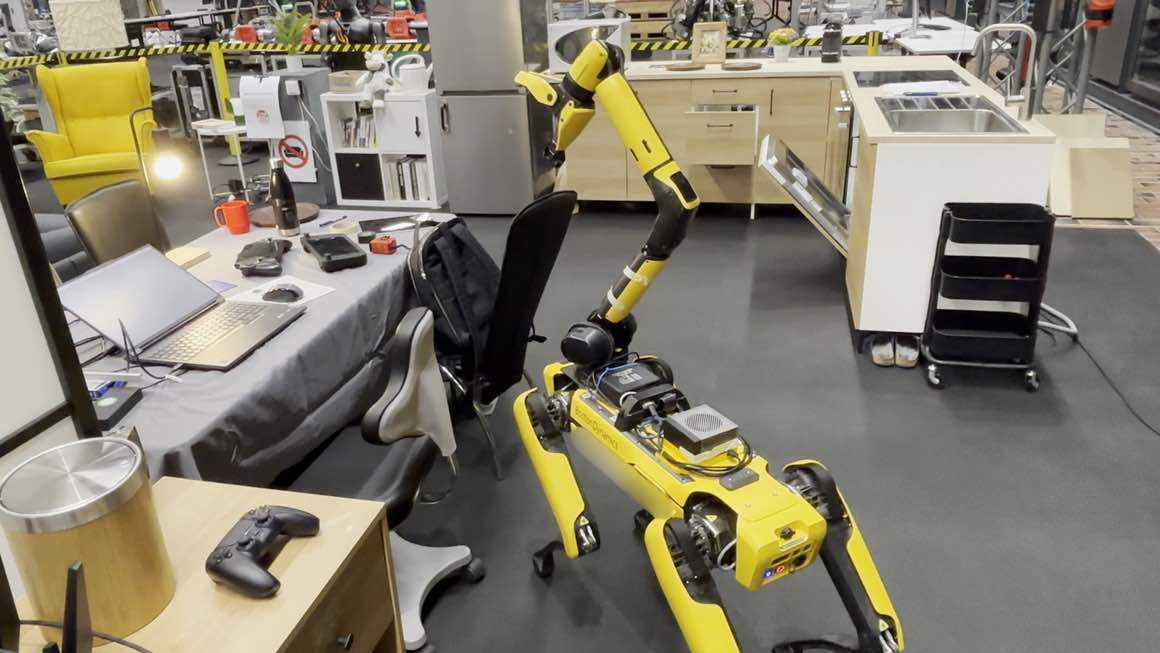} &
    \includegraphics[width=0.32\columnwidth]{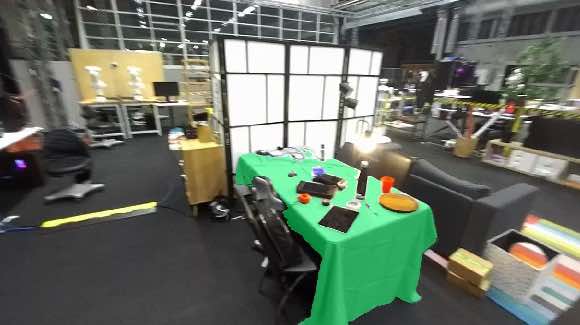} &
    \includegraphics[width=0.32\columnwidth]{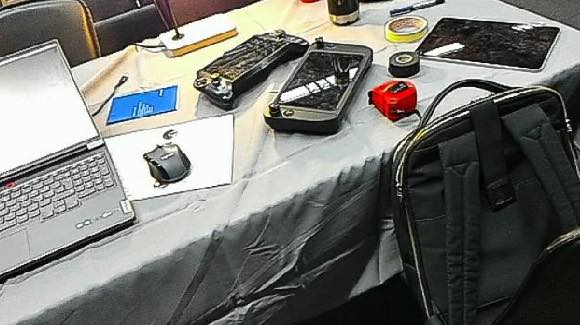} \\[1pt]
    \multicolumn{3}{@{}c@{}}{\scriptsize\textit{Table for four}} \\[3pt]
    \includegraphics[width=0.32\columnwidth]{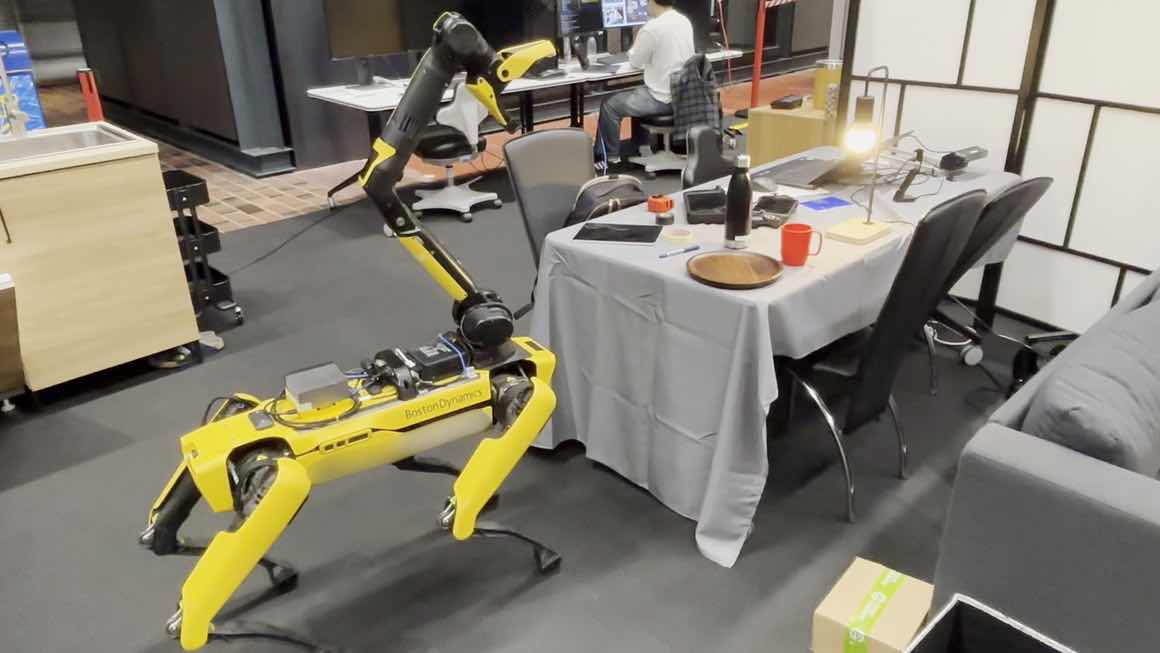} &
    \includegraphics[width=0.32\columnwidth]{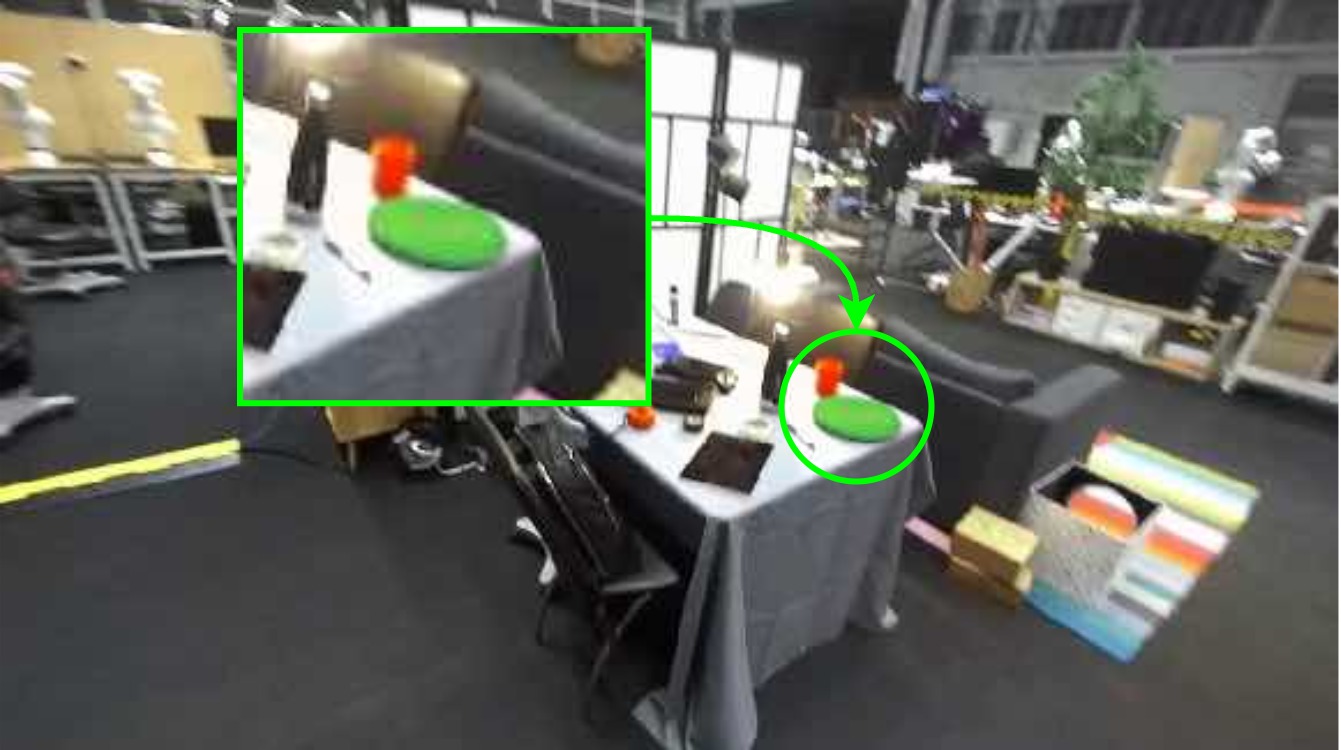} &
    \includegraphics[width=0.32\columnwidth]{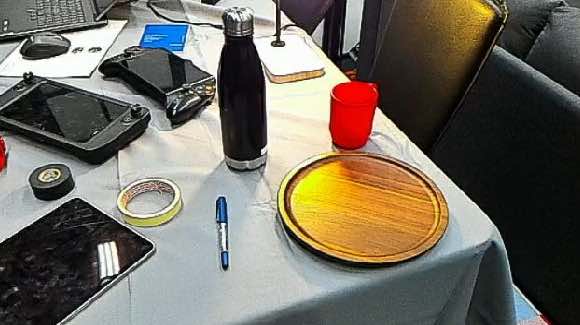} \\[1pt]
    \multicolumn{3}{@{}c@{}}{\scriptsize\textit{Plate on dining table}} \\[3pt]
    \includegraphics[width=0.32\columnwidth]{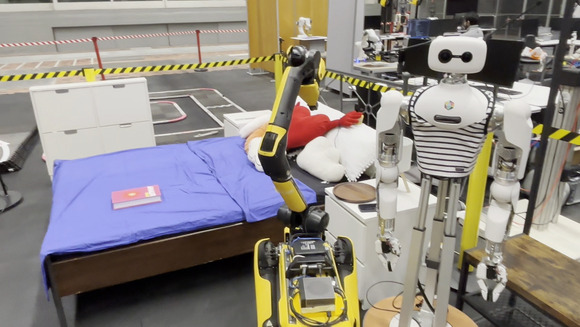} &
    \includegraphics[width=0.32\columnwidth]{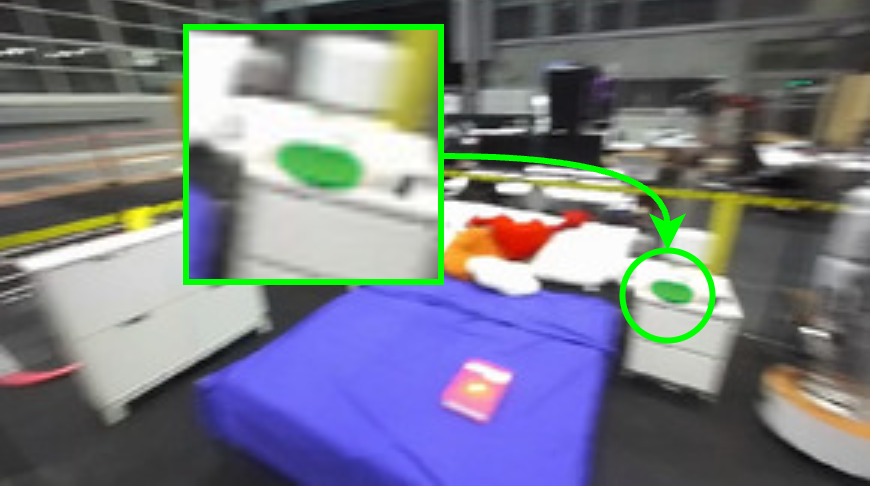} &
    \includegraphics[width=0.32\columnwidth]{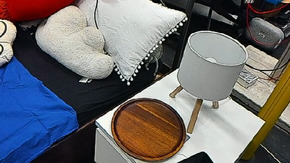} \\[1pt]
    \multicolumn{3}{@{}c@{}}{\scriptsize\textit{Plate on nightstand}} \\[3pt]
    \includegraphics[width=0.32\columnwidth]{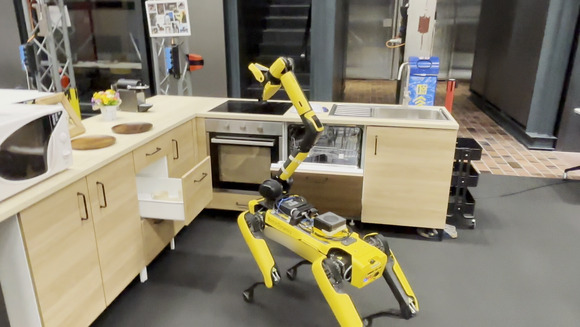} &
    \includegraphics[width=0.32\columnwidth]{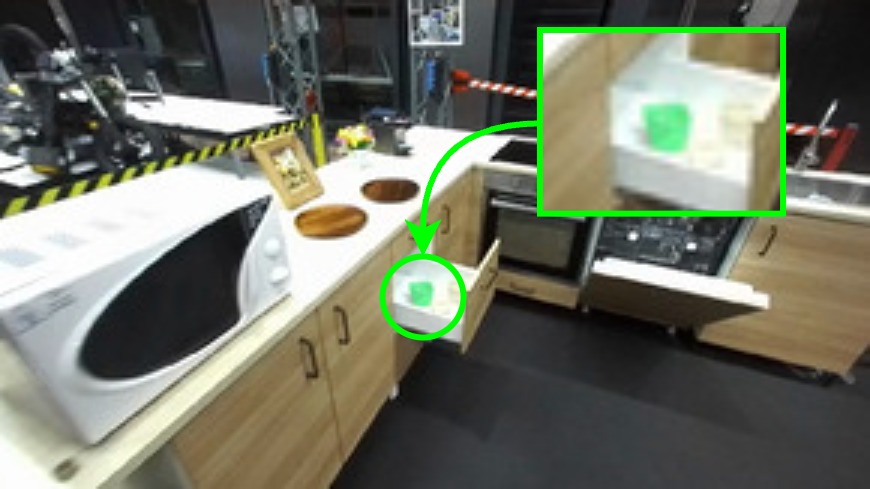} &
    \includegraphics[width=0.32\columnwidth]{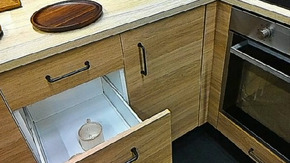} \\[1pt]
    \multicolumn{3}{@{}c@{}}{\scriptsize\textit{Cup in the drawer}} \\
    \end{tabular}
    \caption{\textbf{Real-world robot navigation.} Spot navigates to queried objects in an iPad-scanned indoor scene. \textbf{(a)}~Third-person view of the robot at the goal. \textbf{(b)}~Top-1 retrieved image with segmentation mask ({\color{green}green}). \textbf{(c)}~Spot's onboard camera view upon arrival.}
    \label{fig:robotqualitative}
\end{figure}

\begin{figure}[htbp]
    \centering
    \setlength{\fboxsep}{0pt}
    \setlength{\fboxrule}{1.5pt}

    {\small \textsc{Without VLM Re-Rank}} \\ \vspace{0.3em}
    \begin{minipage}{0.32\linewidth}
        \centering
        \fcolorbox{green}{white}{\includegraphics[width=\linewidth]{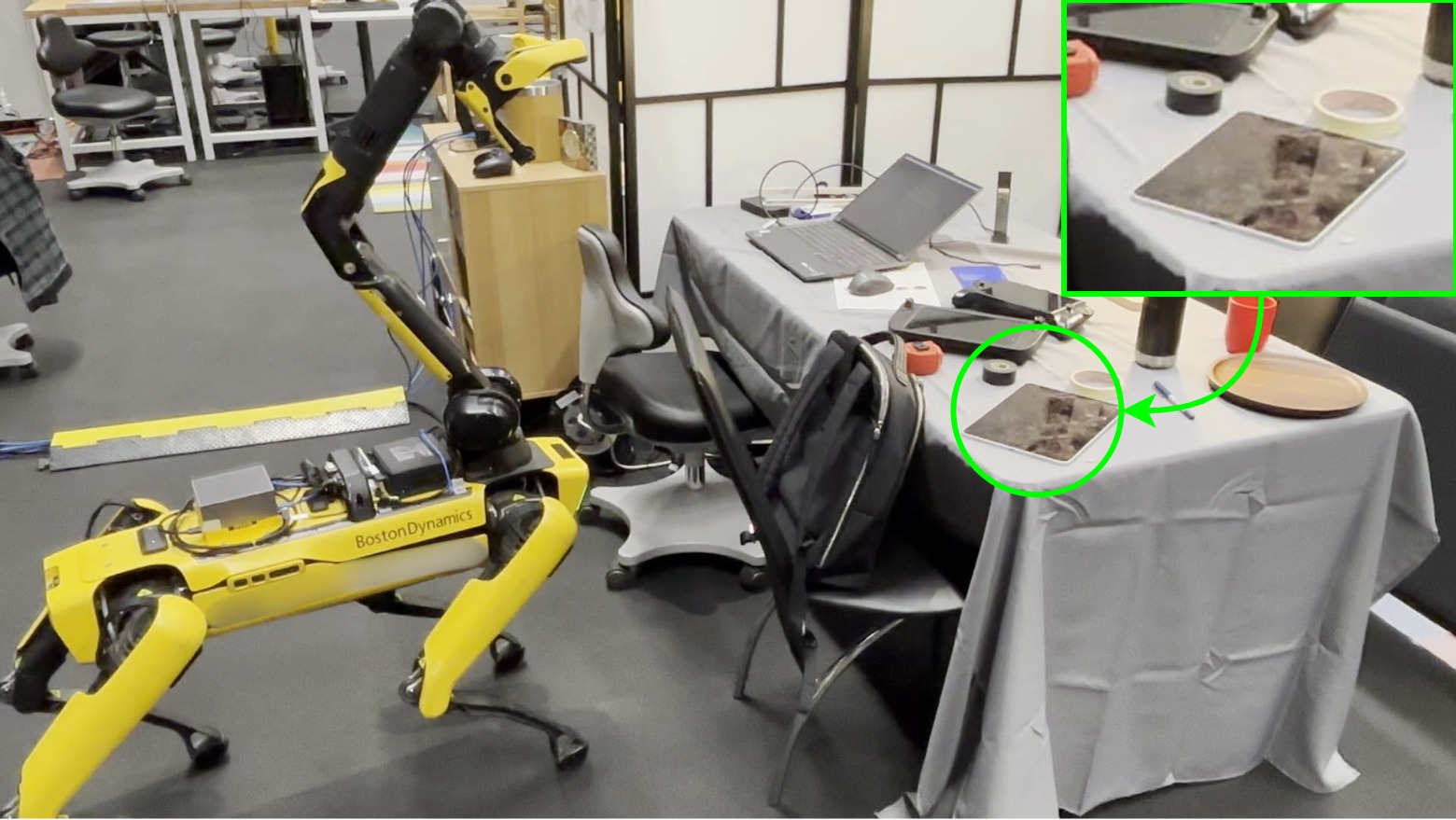}} \\
        \scriptsize\textit{iPad on the dining table}
    \end{minipage}
    \hfill
    \begin{minipage}{0.32\linewidth}
        \centering
        \fcolorbox{red}{white}{\includegraphics[width=\linewidth]{figures/ablationROBOT/ipad.jpg}} \\
        \scriptsize\textit{iPhone on the table}
        \vspace{0.2em}
    \end{minipage}
    \hfill
    \begin{minipage}{0.32\linewidth}
        \centering
        \fcolorbox{red}{white}{\includegraphics[width=\linewidth]{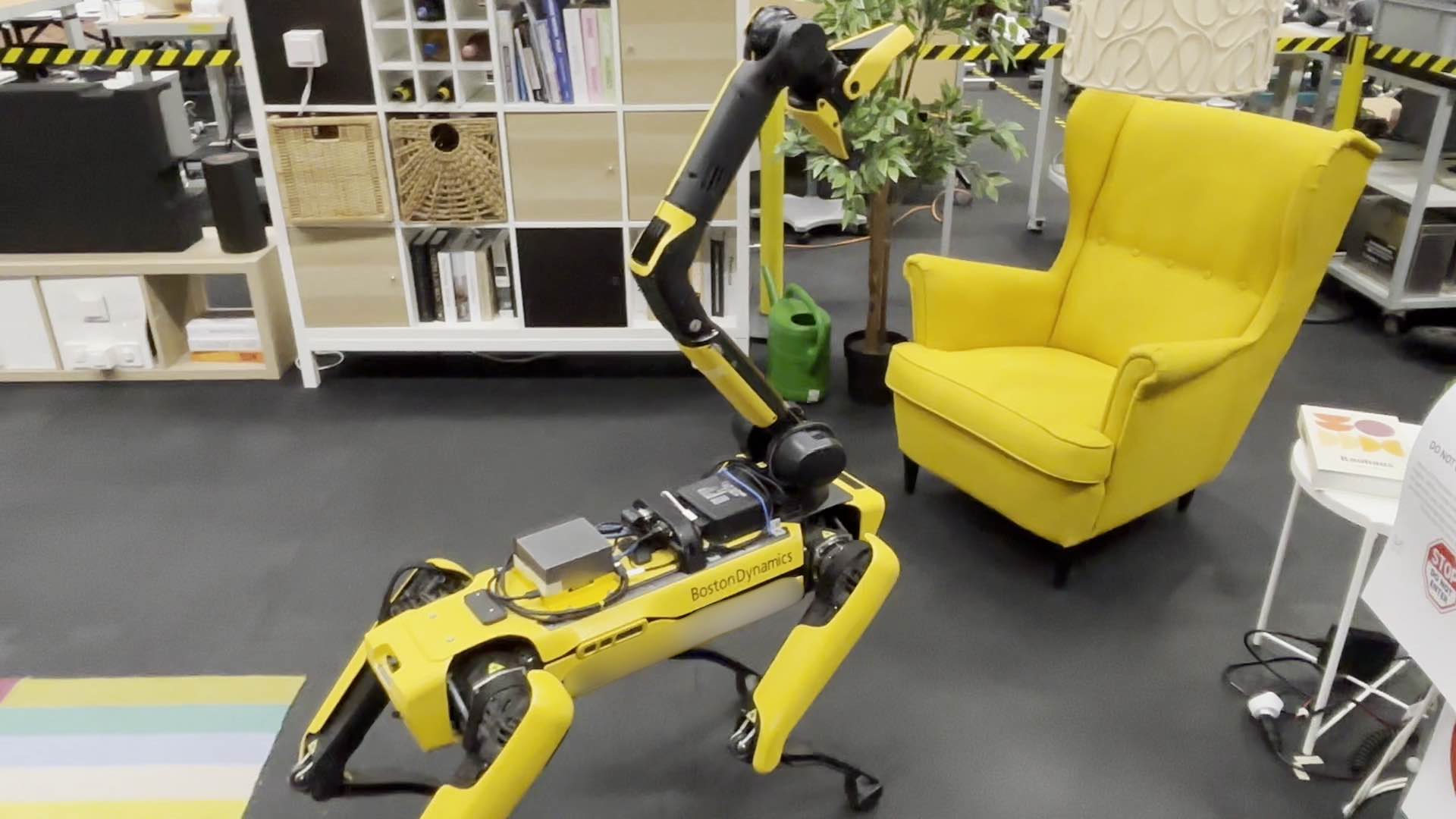}} \\
        \scriptsize\textit{Lit lamp}
    \end{minipage}

    \vspace{0.3em}

    {\small \textsc{With VLM Re-Rank}} \\ \vspace{0.3em}
    \begin{minipage}{0.32\linewidth}
    \centering
    \fcolorbox{green}{white}{\includegraphics[width=\linewidth]{figures/ablationROBOT/ipad.jpg}} \\
    \scriptsize\textit{iPad on the dining table}
    \end{minipage}
    \hfill
    \begin{minipage}{0.32\linewidth}
        \centering
        \fcolorbox{green}{white}{\includegraphics[width=\linewidth]{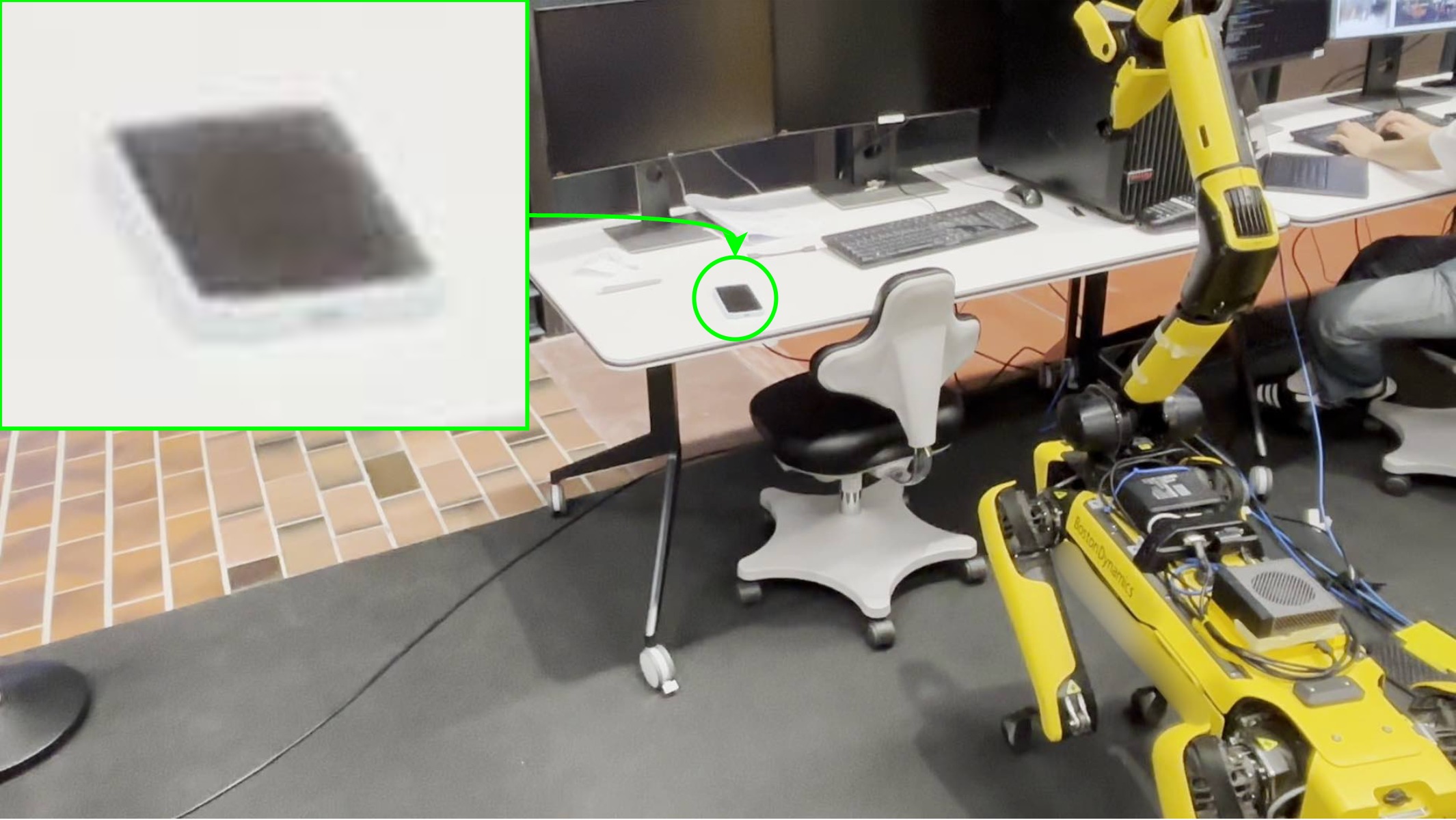}} \\
        \scriptsize\textit{iPhone on the table}
        \vspace{0.2em}
    \end{minipage}
    \hfill
    \begin{minipage}{0.32\linewidth}
        \centering
        \fcolorbox{green}{white}{\includegraphics[width=\linewidth]{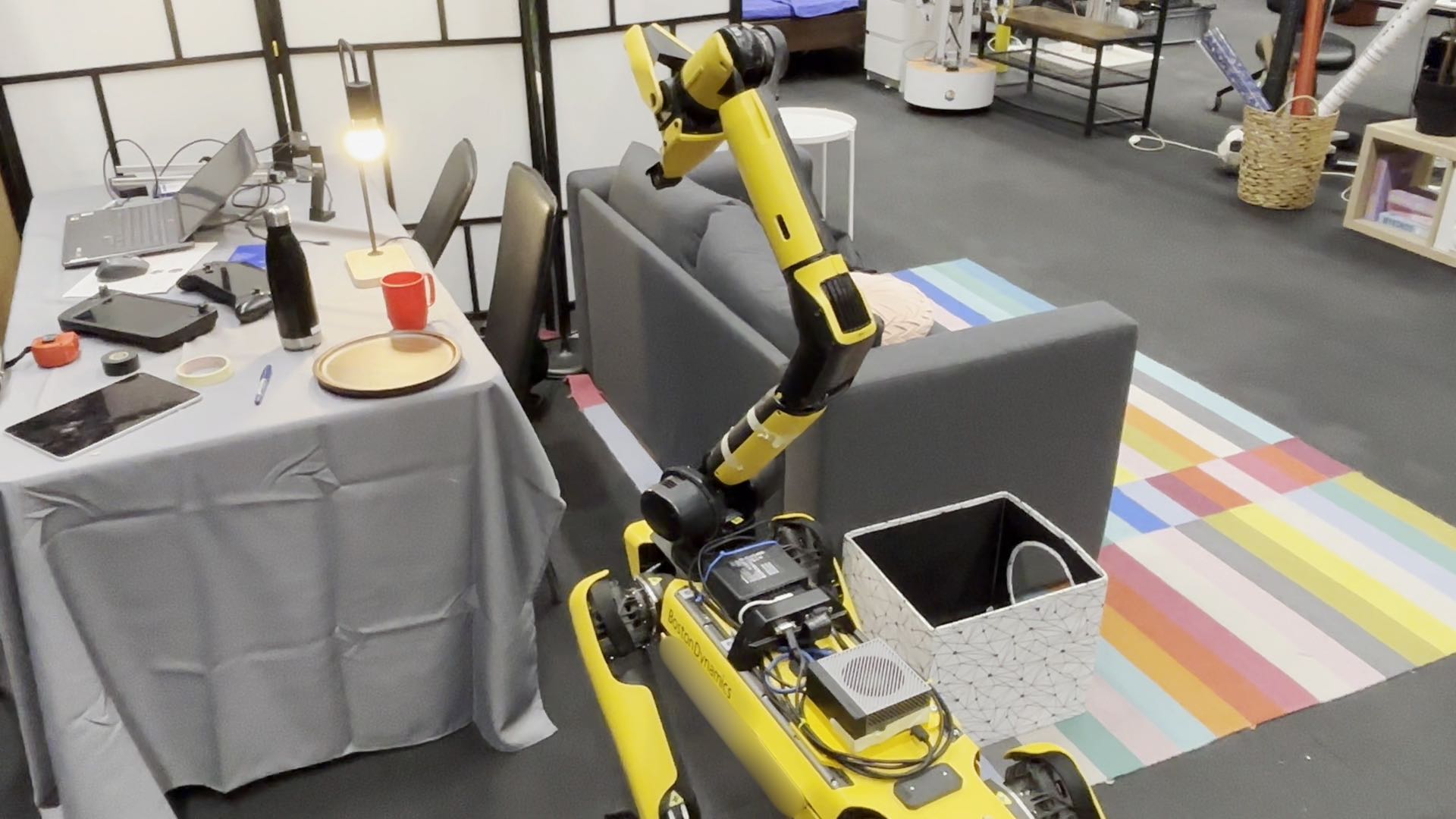}} \\
        \scriptsize\textit{Lit lamp}
    \end{minipage}

    \caption{\textbf{VLM re-ranking ablation.} SigLIP2 retrieves semantically similar but incorrect views (\textcolor{red}{red}); VLM re-ranking corrects all three queries (\textcolor{green}{green}).}
    \label{fig:ablationVLMvsSiglip}
\end{figure}

\subsection{Real-World Experiments}
We deploy our pipeline on a Boston Dynamics Spot robot: the scene is pre-scanned with an iPad, keyframes indexed offline, and the robot navigates to queried objects (\Cref{fig:robotqualitative}). The evaluation included \textit{spatial} reasoning (e.g.: `plate on nightstand'), \textit{fine-grained} object recognition (`the red mug') and \textit{concept fulfillment} (`table for four'). Over 25 trials, our full pipeline achieves 88.0\% SR (22/25) vs.\ 76.0\% (19/25) for the embedding-only variant, a +12\% gain from VLM re-ranking. \Cref{fig:ablationVLMvsSiglip} illustrates how re-ranking corrects fine-grained errors that embedding retrieval alone misses.

\section{CONCLUSION}
We presented a lightweight pipeline for open-vocabulary 3D object localization and navigation that operates directly on posed RGB-D keyframes, avoiding the need for explicit 3D scene reconstruction such as point clouds, voxel grids, or scene graphs. By combining fast embedding retrieval with selective VLM re-ranking, text-prompted segmentation with multi-view spatial fusion, and goal-directed navigation, our approach enables efficient target localization directly from image memory. Despite its simplicity and training-free design, the method demonstrates strong performance across multiple benchmarks and scenarios, highlighting the effectiveness of reasoning directly over image observations rather than constructing dense intermediate representations.

While our formulation focuses on object-centric localization from individual views, i mage by a suitable amount.






\bibliographystyle{IEEEtran}
\bibliography{root
}

\end{document}